\documentclass[journal]{IEEEtran}
\usepackage{subfig}
\usepackage{amsmath}
\usepackage{float}
\usepackage{multirow}
\usepackage{amsmath,amsfonts,amssymb}
\usepackage{algorithmic,algorithm}
\usepackage{pifont,circuitikz}
\usepackage{anysize}
\usepackage{algorithm}
\algsetup{linenosize=\footnotesize}
\usepackage{array}
\usepackage{cite}
\usepackage{booktabs}
\usepackage[strings]{underscore}
\usepackage{tabu}
\usepackage{longtable}
\usepackage{pdflscape}  
\usepackage{longtable}
\usepackage{csquotes}
\usepackage{url}
\ifCLASSINFOpdf
\else
\fi
\begin{document}
%
\title{Improvement over  Pinball Loss Support Vector Machine}
%
%
%

\author{Pritam Anand,~
        Reshma Rastogi
        and~ Suresh Chandra.~
\thanks{ Pritam Anand is with the Dhirubhai Ambani Institute of Information Technology,
	Gandhinagar Gujrat, India – 382007.
	 (e-mail: ltpritamanand@gmail.com, pritam$\_$anand@daiict.ac.in)}
\thanks{  Reshma Rastogi is with the Department of Computer Science, South Asian University, New Delhi-110021. (e-mail: reshma.khemchandani@sau.ac.in)}
\thanks{ Suresh Chandra was with the Department of Mathematics, Indian Institute of Technology, Delhi-110016. (e-mail: sureshiitdelhi@gmail.com.)}
\thanks{}}

%
%

\markboth{Preprint }%
{Shell \MakeLowercase{\textit{et al.}}: }
%



\maketitle

%

\begin{abstract}
Recently, there have been several papers that discuss the extension of the Pinball loss Support Vector Machine (Pin-SVM) model, originally proposed by Huang et al.,\cite{pinsvmpath}\cite{pinsvm}. Pin-SVM classifier deals with the pinball loss function, which has been defined in terms of the parameter $\tau$.  The parameter $\tau$ can take values in $[ -1,1]$. The existing Pin-SVM model requires to solve the same optimization problem for all values of $\tau$ in $[ -1,1]$. In this paper, we improve the existing Pin-SVM model for the binary classification task. At first, we note that there is major difficulty in Pin-SVM model (Huang et al. \cite{pinsvmpath}) for $ -1 \leq \tau < 0$. Specifically, we show that the Pin-SVM model requires the solution of different optimization problem for $ -1 \leq \tau < 0$. We further propose a unified model termed as  Unified Pin-SVM which results in a QPP valid for all $-1\leq \tau \leq 1$ and hence more convenient to use.
The proposed Unified Pin-SVM model can obtain a significant improvement in accuracy over the existing Pin-SVM model which has also been empirically justified by extensive numerical experiments with real-world datasets.
	
\end{abstract}


\section{Introduction}
Support Vector Machines (SVMs)\cite{VapnikCortes}\cite{statistical_learning_theory}\cite{GUNNSVM} are popular machine learning algorithms.  These algorithms are based on Structural Risk Minimization (SRM) principle\cite{statistical_learning_theory}. For binary classification problem  with given training set $ T= \{(x_i,y_i):x_i \in \mathbf{R}^n, y_i\in \{ -1,1\}, i=1.2,...,\textit{l}  \}$, SVM models obtain a separating kernel generated decision function $w^T\phi(x_i)+b=0$  by minimizing a good trade-off between the empirical risk and model complexity in its optimization problem. SVM models use a loss function to measure the empirical risk of the given training set. For minimizing the model complexity, SVM models minimize a regularization term in their optimization problem. 

The standard  C-SVM model minimizes the Hinge loss function along with the $L_2$-norm regularization in its formulation. Thus, it minimizes
\begin{equation}
\frac{1}{2} ||w||_2^2+ C_0\sum_{i=1}^{l} L_{Hinge}(1-y_i(w^T\phi(x_i)+b)),
\end{equation}
where $L_{Hinge} = max(u,0)$ is the Hinge loss function and $C_0 \geq 0$ is the user supplied parameter.  The use of Hinge loss function in C-SVM model makes it ignore the data points which satisfy $y_i(w^T\phi(x_i)+b) > 1$. There are few data points satisfying  $y_i(w^T\phi(x_i)+b) \leq 1$, which contribute for the empirical risk. These data points are called `support vectors' and lie near the boundary of the separating hyperplane $w^T\phi(x)+b=0$.  The separating hyperplane in C-SVM model is only constructed by using these support vectors. This causes the sparsity in C-SVM model. But, data points near the boundary of the separating hyperplane may be noisy which can mislead the resulting separating hyperplane.
To improve the C-SVM model, Huang et al. \cite{pinsvm} have suggested to use the pinball loss function\cite{quantile1} in SVM model. For the classification problem, the pinball loss function is given by
{\small \begin{equation}\label{eqn2}
L^{\tau \geq 0}_{pin}(u)  = 
\begin{cases} 
u, ~~  \mbox{if}~~ u \geq 0,\\
-\tau u, otherwise,
\end{cases} 	
\end{equation}}
where $ 0 \leq \tau \leq 1 $ is its parameter. For $\tau=0$, the pinball loss function reduces to the Hinge loss function. For $\tau=1$, it reduces to the $l_1$ loss function.                  

The Pin-SVM model (Huang et al.,\cite{pinsvm}) minimizes the empirical risk using the pinball loss function along with the $L_2$-norm regularization in its formulation. This leads to the following optimization problem
\begin{eqnarray}
\min_{(w,b)} \frac{1}{2}||w||_2^2 + C_0\sum_{i=1}^{l} L^{\tau \geq 0}_{pin}(1-y_i(w^T\phi(x_i)+b)), \label{pin_svm-positive_tau}  \nonumber
\end{eqnarray}
which can be equivalently converted to the following optimization problem 
\begin{eqnarray}
\min_{(w,b,\xi)} \frac{1}{2}||w||_2^2 + C_0 \sum_{i=1}^{l} \xi_i \nonumber\\
& \hspace{-72mm} \mbox{subject to,}\nonumber\\
& \hspace{-40mm} y_i(w^T\phi(x_i)+b) \geq  1- \xi_i  ,\nonumber\\
&  \hspace{-40mm}  y_i(w^T\phi(x_i)+b) \leq 1+\frac{\xi_i}{\tau},
\label{pinsvm1}
\end{eqnarray}
where $\xi_1,\xi_2,..,\xi_l$ are slack variables and  $C_0,~0 \leq \tau \leq 1$ are user supplied parameters.

The pinball loss function in Pin-SVM model penalizes (assigns positive risk) every data point but, with different rate. Data points satisfying $y_i(w^T\phi(x_i)+b) \leq 1 $ are penalized with unit rate and other data points are penalized with comparatively lower rate $\tau$. This penalization in Pin-SVM model causes it to also minimize the scatter of data points along the separating hyperplane. But then, it takes away the very nice property of SVM, namely sparsity. However, the Pin-SVM model is a general SVM model in the sense that it can reduce to the standard C-SVM model for its parameter $\tau=0$.

 To reduce the effect of the unbalanced class labeling, we consider a $l$-dimensional vector $C=(C_1,C_2,\ldots C_l)$, rather than a single constant $C_0$, such that\\
\begin{equation}
C_i= 
\begin{cases}
C_0,~~~~~~~~~~~~  y_i = + 1 ,\\
p C_0 ,~~~~~~~~~~~~ y_i = -1,
\end{cases} \label{ci}
\end{equation}

 where $p$ is defined as \\
 
 $p=~\frac{\mbox{number of data points on `Class +1'}}{\mbox{number of data points in `Class -1'}}$ 

 and seek the solution of following optimization problem 
\begin{eqnarray}
\min_{(w,b,\xi)} \frac{1}{2}||w||_2^2 +  \sum_{i=1}^{l}C_i\xi_i \nonumber\\
& \hspace{-80mm} \mbox{subject to,}\nonumber\\
& \hspace{-40mm} y_i(w^T\phi(x_i)+b) \geq  1- \xi_i  ,\nonumber\\
&  \hspace{-30mm}  y_i(w^T\phi(x_i)+b) \leq 1+\frac{\xi_i}{\tau}, \tau \geq 0.
\label{pinsvm2}
\end{eqnarray}
  
   Rather than solving the primal problem (\ref{pinsvm2}), we prefer to solve its Wolfe's dual problem, which is obtained as follows
   \begin{eqnarray}
   \min_{(\alpha, \beta)} \frac{1}{2} \sum_{i=1}^{l}\sum_{j=1}^{l}(\alpha_j-\beta_j)(\alpha_i-\beta_i)y_iy_jK(x_i,x_j) \nonumber \\
    & \hspace{-110mm}-\sum\limits_{i=1}^{l}(\alpha_i - \beta_i) \nonumber \\
   & \hspace{-130mm} \mbox{subject to,} \nonumber \\
   & \hspace{-102mm} \sum\limits_{i=1}^{l}(\alpha_i-\beta_i)y_i = 0, \nonumber\\
   & \hspace{-75mm}  C_i-\alpha_i -\frac{1}{\tau}\beta_i = 0,~~ for ~i=1,2,..,l, \nonumber \\
   & \hspace{-80mm} \alpha_i \geq 0,~~ \beta_i \geq 0,~~ for ~i=1,2,..,l. \label{dualpinsvm}
   \end{eqnarray}
   More information about properties of pinball loss function and Pin-SVM model can be found in (Huang et al.,\cite{pinsvm}).
   
 We organize the rest of this paper as follows. Section \ref{sec1} describes the optimization problem of Pin-SVM model for $ -1 \leq \tau < 0$ as proposed in (Hunag et al., \cite{pinsvmpath}).
   In section \ref{sec2}, we  derive the right optimization problem for Pin-SVM model for $ -1 \leq \tau < 0$. In section \ref{sec3}, we propose a unified optimization problem which can obtain the solution of Pin-SVM model without bothering the sign of its parameter $\tau$ in $[-1,1]$. We term this proposed model as Unified Pin-SVM model. Section \ref{sec4} presents numerical results which empirically verify that the proposed Unified Pin-SVM model corrects the existing Pin-SVM model by minimizing the pinball loss function in true sense.  
\section{Pinball loss function with negative $\tau$ value and SVM model}\label{sec1}
 Huang et al. extended the pin-ball loss function for negative $\tau$ using the same expression in their work (Huang et al. \cite{pinsvmpath}) . The pinball loss function with negative $\tau$ is given by
{\small \begin{equation}
L^{\tau\leq 0}_{pin}(u) = 
\begin{cases} 
u, ~~ ~~  ~~\mbox{if}~~ u \geq 0.\\
-\tau u, ~~ othewise. \\
\end{cases} 
\label{pinball_neg_tau}
\end{equation}}
The  above pinball loss function (\ref{pinball_neg_tau}) is convex loss function for   $  \tau \geq -1 $.
%
%
 Huang et al. have formulated the Pin-SVM model for  $ -1 \leq \tau < 0$ using
\begin{eqnarray}
\min_{(w,b)} \frac{1}{2}||w||_2^2 + C\sum_{i=1}^{l} L^{\tau \leq 0}_{pin}(1-y_i(w^T\phi(x_i)+b)). \label{pinsvm_with_neg}
\end{eqnarray}

For minimizing (\ref{pinsvm_with_neg}), they have chosen to minimize the following Quadratic Programming Problem (QPP)
\begin{eqnarray}
\min_{(w,b,\xi)} \frac{1}{2}||w||_2^2 +  \sum_{i=1}^{l}C_i\xi_i \nonumber\\
& \hspace{-74mm} \mbox{subject to,}\nonumber\\
& \hspace{-35mm} y_i(w^T\phi(x_i)+b) \geq  1- \xi_i  ,\nonumber\\
&  \hspace{-32mm}  y_i(w^T\phi(x_i)+b) \leq 1+\frac{\xi_i}{\tau},\label{opt_for_pin_neg}
\end{eqnarray}
  where $ -1 \leq \tau < 0$ is user supplied parameter.
It should be noted that, Huang et al. have used the same Pin-SVM optimization problem for both  positive and negative values of $\tau$ in $[-1,1]$. Contrary to this, we claim in the next section of this paper that the Pin SVM model for $ -1 \leq \tau < 0$ requires the solution of a QPP which is different from (\ref{opt_for_pin_neg}). 
\section{Pin-SVM with negative $\tau$ values}\label{sec2}
This paper improves the existing Pin-SVM model for $-1 \leq \tau < 0$ (Huang et al.,\cite{pinsvmpath}).  We shall show that the optimization problem of existing Pin-SVM model for $-1 \leq \tau < 0$ obtained in (Huang et al.,\cite{pinsvmpath}) is not correct and  derive the right optimization problem for it.

The pinball loss function (\ref{pinball_neg_tau}) has been used in (Huang et al.,\cite{pinsvmpath}\cite{pinsvm}) for $ -1\leq \tau \leq 1$.  
At first, we consider the  loss function
 \begin{equation}
L_{pin}^{\tau}(u) = max( u, -\tau u)  
\label{mpinball_neg_tau}
\end{equation}
 for $ -1 \leq \tau \leq 1$.  For $-1\leq \tau \leq  1$, we can obtain
$max( u, -\tau u) = 
\begin{cases} 
u, ~~ ~\mbox{if}~ u \geq 0.\\
-\tau u, \mbox{ otherwise}.
\end{cases} $\\ It makes us realize that the pinball loss function is equivalent to the $max( u, -\tau u)$ for $ -1 \leq \tau \leq 1$. 


Now, we shall state and justify our claim about the existing Pin-SVM model with $ -1 \leq \tau < 0$. We claim that the Pin-SVM model with $ -1 \leq \tau < 0$  (problem (\ref{pinsvm_with_neg})) is not equivalent to the solving QPP (\ref{opt_for_pin_neg}) used in (Huang et al., \cite{pinsvmpath}) and vice-versa. The justification of this claim is detailed as follows.

The Pin-SVM for  $ -1 \leq \tau < 0$ (problem (\ref{pinsvm_with_neg})) is equivalent to

\begin{eqnarray}
 \min_{(w,b)} \frac{1}{2}||w||_2^2 + C_0\sum_{i=1}^{l} max((1-y_i(w^T\phi(x_i)+b)), \nonumber\\
-\tau(1-y_i(w^T\phi(x_i)+b))) \mbox{~where~} -1 \leq \tau < 0. \label{pinsvm_opt1_neg}
\end{eqnarray}

Let us consider  slack variables $\xi_i =max((1-y_i(w^T\phi(x_i)+b)), -\tau(1-y_i(w^T\phi(x_i)+b))) ,i=1,2,...,\textit{l}$. Then, the optimization problem (\ref{pinsvm_opt1_neg}) of Pin-SVM can be given by
\begin{eqnarray}
\min_{(w,b,\xi)} \frac{1}{2}||w||_2^2 +  C_0\sum_{i=1}^{l} \xi_i \nonumber\\
& \hspace{-95mm} \mbox{subject to,}\nonumber\\
& \hspace{-65mm}\xi_i \geq 1- y_i(w^T\phi(x_i)+b) ,\nonumber\\
&  \hspace{-38mm}\xi_i \geq -\tau(1- y_i(w^T\phi(x_i)+b)), -1 \leq \tau < 0. \label{pinsvm_neg_opt2}
\end{eqnarray}
Since $\tau < 0 $ in above optimization problem (\ref{pinsvm_neg_opt2}), so its second constraint \begin{eqnarray}
\xi_i \geq -\tau(1- y_i(w^T\phi(x_i)+b)) \mbox{~is equivalent to~}  \nonumber\\   & \hspace{-110mm} y_i(w^T\phi(x_i)+b) \geq 1 + \frac{\xi_i}{\tau}.  \nonumber\\
\mbox {Similarly, the first constraint of problem (\ref{pinsvm_neg_opt2}) ~~}\nonumber \\  \xi_i \geq 1- y_i(w^T\phi(x_i)+b) \mbox{~is equivalent to~}  \nonumber  \\ &\hspace{-90mm} y_i(w^T\phi(x_i)+b) \geq 1- \xi_i.   
\end{eqnarray}
Now, optimization problem (\ref{pinsvm_neg_opt2}) can be obtained as 
\begin{eqnarray}
\min_{(w,b,\xi)} \frac{1}{2}||w||_2^2 + C_0 \sum_{i=1}^{l} \xi_i \nonumber\\
& \hspace{-70mm} \mbox{subject to,}\nonumber\\
& \hspace{-45mm} y_i(w^T\phi(x_i)+b) \geq  1- \xi_i  ,\nonumber\\
&  \hspace{-45mm}  y_i(w^T\phi(x_i)+b) \geq 1+\frac{\xi_i}{\tau}, \label{actual_opt_pin_neg}
\end{eqnarray}
where $-1 \leq \tau < 0$,
which is different from QPP (\ref{opt_for_pin_neg}) used in (Huang et al., \cite{pinsvmpath}). It also infers that QPP (\ref{actual_opt_pin_neg}) is the actual minimizer of the Pin-SVM model with $ -1 \leq \tau < 0$ (problem (\ref{pinsvm_with_neg})). 

\subsection{Solution of QPP for Pin-SVM with negative $\tau$ }
For unbalanced training set, the Pin-SVM optimization problem with $ -1 \leq \tau < 0$  can also be modified as 
\begin{eqnarray}
\min_{(w,b,\xi)} \frac{1}{2}||w||_2^2 + \sum_{i=1}^{l}C_i \xi_i \nonumber\\
& \hspace{-70mm} \mbox{subject to,}\nonumber\\
& \hspace{-45mm} y_i(w^T\phi(x_i)+b) \geq  1- \xi_i  ,\nonumber\\
&  \hspace{-45mm}  y_i(w^T\phi(x_i)+b) \geq 1+\frac{\xi}{\tau},  \label{actual_opt_pin_neg1}
\end{eqnarray} 
where $C_i$ are as defined in (\ref{ci}) and $ -1\leq \tau < 0$. 
In order to find the solution of above primal problem, we need to derive its corresponding Wolfe`s dual problem. For this, we construct the Lagrangian function for primal problem (\ref{actual_opt_pin_neg1}) as follows
\begin{eqnarray}
L(w,b,\xi_i,\alpha_i,\beta_i)=
\frac{1}{2}||w||_2^2+ C\sum_{i=1}^{l}\xi_i  \nonumber \\ -\sum_{i=1}^{l}{\alpha_i}( y_i(w^T\phi(x_i)+ b)-1+\xi_i) \nonumber\\ -\sum_{i=1}^{l}{\beta_i}( y_i(w^T\phi(x_i)+ b)-1-\frac{\xi_i}{\tau}).
\end{eqnarray}
 We list some relevant Karush-Kuhn-Tucker(KKT) conditions for the optimization problem (\ref{actual_opt_pin_neg1}) as follows
\begin{eqnarray}
& \hspace{-20mm} \frac{\partial L}{\partial w} =  w- \sum \limits_{i=1}^{l}(\alpha_i+\beta_i)y_i\phi(x_i) = 0,\label{r11}\\
&\hspace{-35mm} \frac{\partial L}{\partial b} =  \sum \limits_{i=1}^{l}(\alpha_i+\beta_i)y_i = 0, \\
& \hspace{-15mm}\frac{\partial L}{\partial \xi_i} =  C- \alpha_i +\frac{1}{\tau}\beta_i = 0,
i=1,2,..,l.\label{r1}
\end{eqnarray}
Using the KKT conditions, the Wolfe's dual of the primal problem (\ref{actual_opt_pin_neg1}) can be obtained as follows
\begin{eqnarray}
\min_{(\alpha, \beta)} \frac{1}{2} \sum_{i=1}^{l}\sum_{j=1}^{l}(\alpha_j+\beta_j)(\alpha_i+\beta_i)y_iy_j(\phi(x_i)^T\phi(x_j)) \nonumber \\
 &  \hspace{-120mm}-\sum \limits _{i=1}^{l}(\alpha_i + \beta_i) \nonumber \\
& \hspace{-130mm} \mbox{subject to,} \nonumber \\
& \hspace{-100mm} \sum \limits_{i=1}^{l}(\alpha_i+\beta_i)y_i = 0, \nonumber\\
& \hspace{-105mm}  C-\alpha_i + \frac{1}{\tau}\beta_i = 0, \nonumber \\
& \hspace{-90mm} \alpha_i \geq 0,~~ \beta_i \geq 0, i=1,2,..,l. \label{dualrpsvm} \nonumber
\end{eqnarray}
By using a positive semi-definite kernel $K(x_i,x_j)= \phi(x_i)^T\phi(x_j)$, satisfying  Mercer condition (Mercer,\cite{mercer1909xvi}), the above dual problem  can be obtained as 
\begin{eqnarray}
\min_{(\alpha, \beta)} \frac{1}{2} \sum_{i=1}^{l}\sum_{j=1}^{l}(\alpha_j+\beta_j)(\alpha_i+\beta_i)y_iy_jK(x_i,x_j) \nonumber \\
 &  \hspace{-100mm}-\sum\limits_{i=1}^{l}(\alpha_i + \beta_i) \nonumber \\
& \hspace{-130mm} \mbox{subject to,} \nonumber \\
& \hspace{-105mm} \sum \limits_{i=1}^{l}(\alpha_i+\beta_i)y_i = 0, \nonumber\\
& \hspace{-105mm}  C-\alpha_i + \frac{1}{\tau}\beta_i = 0, \nonumber \\
& \hspace{-90mm} \alpha_i \geq 0,~~ \beta_i \geq 0,~~ i=1,2,..,l. \label{dualrpsvm1}
\end{eqnarray}

After obtaining the solution of the dual problem (\ref{dualrpsvm1}), the value of $w$ can be obtained from the KKT condition (\ref{r11}) as follows
\begin{eqnarray}
w= \sum_{i=1}^{l}(\alpha_i+\beta_i)y_i\phi(x_i).
\end{eqnarray}         
Let us now define the following set\\
$~~~~~~~~~~~~~~~~~ S= \{i:\alpha_{i} > 0 , \beta_{i} > 0   \}$,\\
 Using the complementary slackness condition, we compute the values of the $b$ for each $i \in S$, from
{\small \begin{equation}
b= y_i-w^T\phi(x_i) = y_i-\sum_{j=1}^{l}(\alpha_{j}+\beta_{j})y_jK(x_j,x_i) \label{eq3} 
\end{equation}}
 and take their average value as the final value of the  bias $b$.
For given test point $x \in {R}^n$,  the decision function is obtained as
\begin{eqnarray}
f(x)  =sign( w^T\phi(x)+ b) \nonumber \\
& \hspace{-40mm}  = sign ~(\sum\limits_{j=1}^{l}(\alpha_{j}+\beta_{j})y_jK(x_j,x)+ b).
\end{eqnarray}
\section{A unified QPP for solving Pin-SVM problem}\label{sec3}
We can observe that minimizing the Pin-SVM problem with positive and negative $\tau$ value in $[-1,1]$  results into two different QPPs. Minimizing different QPPs for negative and positive $\tau$ value in Pin-SVM problem may not be handful for searching best $\tau \in [-1,1]$, which corresponds to the optimal accuracy. Taking motivation from this, we also propose a unified optimization problem which can obtain the solution of Pin-SVM problem without bothering about the sign of its parameter  $\tau$.  For a given $\tau \in [-1,1]$, the Pin-SVM model should minimize
\begin{eqnarray}
\min_{(w,b)} \frac{1}{2}||w||_2^2+ C_0\sum_{i=1}^{l}(max(1-y_i(w^T\phi(x_i)+b) \nonumber \\,-\tau(1-y_i(w^T\phi(x_i)+b))\nonumber \label{unified_opt1}.
\end{eqnarray}
After introducing the slack variable $ \xi_i=max(1-y_i(w^T\phi(x_i)+b), -\tau(1-y_i(w^T\phi(x_i)+b))$, the Pin-SVM problem becomes 
\begin{eqnarray}
\min_{(w,b,\xi)} \frac{1}{2}||w||_2^2 + C_0\sum_{i=1}^{l} \xi_i \nonumber\\
& \hspace{-70mm} \mbox{subject to,}\nonumber\\
& \hspace{-45mm}\xi_i \geq 1- y_i(w^T\phi(x_i)+b) ,\nonumber\\
&  \hspace{-40mm}\xi_i \geq -\tau(1- y_i(w^T\phi(x_i)+b)). \label{u_pinsvm_opt3}
\end{eqnarray}
For the unbalanced training set, the suitable Pin-SVM problem can be given by 
\begin{eqnarray}
\min_{(w,b,\xi)} \frac{1}{2}||w||_2^2 + \sum_{i=1}^{l}C_i \xi_i \nonumber\\
& \hspace{-70mm} \mbox{subject to,}\nonumber\\
& \hspace{-45mm}\xi_i \geq 1- y_i(w^T\phi(x_i)+b) ,\nonumber\\
&  \hspace{-35mm}\xi_i \geq -\tau(1- y_i(w^T\phi(x_i)+b)). \label{u_pinsvm_opt2}
\end{eqnarray}
We obtain the Lagrangian function for the primal problem (\ref{u_pinsvm_opt2}) as follow
\begin{eqnarray}
L(w,b,\xi_i,\alpha_i,\beta_i)=
\frac{1}{2}||w||^2+ \sum_{i=1}^{l}C_i\xi_i \nonumber \\ -\sum_{i=1}^{l}{\alpha_i}( y_i(w^T\phi(x_i)+ b)-1+\xi_i) \nonumber\\ & \hspace{-55mm} - \sum \limits_{i=1}^{l}{\beta_i}( \tau(1-y_i(w^T\phi(x_i)+ b))+\xi_i).
\end{eqnarray}
We list some relevant KKT optimality conditions for the optimization problem (\ref{u_pinsvm_opt2}) as follows.
\begin{eqnarray}
& \hspace{-20mm} \frac{\partial L}{\partial w} =  w- \sum \limits_{i=1}^{l}(\alpha_i-\tau\beta_i)y_i\phi(x_i) = 0,\label{ur11}\\
&\hspace{-35mm} \frac{\partial L}{\partial b} =  \sum \limits_{i=1}^{l}(\alpha_i-\tau\beta_i)y_i = 0, \\
& \hspace{-25mm}\frac{\partial L}{\partial \xi_i} =  C_i- \alpha_i - \beta_i = 0,  ~i=1,2,..,l\label{ur1}  
\end{eqnarray}
Using the KKT optimality conditions, the Wolfe's dual of the primal problem (\ref{u_pinsvm_opt2}) is obtained as follows
\begin{eqnarray}
\min_{(\alpha, \beta)} \frac{1}{2} \sum_{i=1}^{l}\sum_{j=1}^{l}(\alpha_j-\tau\beta_j)(\alpha_i-\tau\beta_i)y_iy_jK(x_i,x_j)
 \nonumber \\   & \hspace{-110mm}-\sum \limits_{i=1}^{l}(\alpha_i -\tau\beta_i) \nonumber \\
& \hspace{-130mm} \mbox{subject to,} \nonumber \\
& \hspace{-105mm} \sum \limits_{i=1}^{l}(\alpha_i-\tau\beta_i)y_i = 0, \nonumber\\
& \hspace{-110mm}  C_i-\alpha_i -\beta_i = 0,\nonumber \\
& \hspace{-90mm} \alpha_i \geq 0,~ \beta_i \geq 0,~~  ~i=1,2,..,l. \label{dual_u_psvm}
\end{eqnarray}

If we consider the  replacement of variable $\beta:=|\tau|\beta$ in  dual problem (\ref{dual_u_psvm}) and define a signum function $s_u = \begin{cases}
1 ~~ \mbox{if}~ u \geq 0,\\
-1 ,~~ otherwise,
\end{cases}$then, the dual problem (\ref{dual_u_psvm}) can be given by

{\small \begin{eqnarray}
\min_{(\alpha, \beta)} \frac{1}{2} \sum_{i=1}^{l}\sum_{j=1}^{l}(\alpha_j-s_\tau\beta_j)(\alpha_i-s_\tau\beta_i)y_iy_jK(x_i,x_j)
\nonumber \\   & \hspace{-110mm}-\sum\limits_{i=1}^{l}(\alpha_i -s_\tau\beta_i) \nonumber \\ 
& \hspace{-130mm} \mbox{subject to,} \nonumber \\
& \hspace{-105mm} \sum\limits_{i=1}^{l}(\alpha_i-s_\tau\beta_i)y_i = 0, \nonumber\\
& \hspace{-110mm}  C_i-\alpha_i -\frac{\beta_i}{|\tau|} = 0, \nonumber \\
& \hspace{-90mm} \alpha_i \geq 0,~ \beta_i \geq 0,~~ i=1,2,..,l. \label{dual_u_psvm1}
\end{eqnarray}}
 It is notable that for  $ 0 \leq \tau \leq 1$, the proposed dual problem (\ref{dual_u_psvm1}) is equivalent to  dual problem (\ref{dualpinsvm}) of Pin-SVM model. For $ -1 \leq \tau < 0$  , the  proposed dual problem (\ref{dual_u_psvm1}) can be found to be equivalent to the dual problem (\ref{dualrpsvm1}) of Pin-SVM model for $ -1 \leq \tau < 0$. This is because $ \tau = s_{\tau}|\tau|$.
 
After obtaining the solution of the dual problem (\ref{dual_u_psvm1}), the value of $w$ can be given by
\begin{eqnarray}
w= \sum_{i=1}^{l}(\alpha_i-s_\tau\beta_i)y_i\phi(x_i).\label{ww}
\end{eqnarray}
 For finding the value of $b$, we consider each index $i$ such that $\alpha_{i} > 0$ and $\beta_{i} > 0 $, and compute the value of $b$ using the complementary slackness condition as follow
{\small  \begin{equation}
 b= y_i-w^T\phi(x_i) = y_i-\sum_{j=1}^{l}(\alpha_{j}-s_\tau\beta_{j})y_jK(x_j,x_i). \label{bb}
 \end{equation}}
  We consider the  final value of  $b$ by taking  the average  over all  possible values of $b$.
 For given test point $x \in {R}^n$,  the decision function is obtained as
 \begin{eqnarray}
 f(x)  =sign( w^T\phi(x)+ b) \nonumber \\
 & \hspace{-40mm}  = sign ~(\sum\limits_{j=1}^{l}(\alpha_{j}-s_\tau\beta_{j})y_jK(x_j,x)+ b).\label{pred}
 \end{eqnarray}
 \begin{algorithm}\label{algo1}
 	\caption{Unified Pin-SVM }
 	\textbf{Input:-} Training set $T =\{(x_i,y_i):x_i \in \mathbb{R}^n, y_i\in \{ -1,1\}, i=1.2,...,\textit{l}  \}$, test data $x \in \mathbb{R}^n$, and parameter $\tau$.\\
 	\textbf{Output:-} Predicted  label for test data $x$.
 	\begin{algorithmic}[1]
 		\item [(i)] Select a penalty parameter $C_0>0$ and kernel parameter $q$, if required . These parameter are commonly selected through validation.   
 		\item [(ii)]For $i=1,2,..,l$, compute $C_i$ using (\ref{ci}). 
 		\item [(iii)] For the linear kernel compute $ k(x_i,x_j)= x_i^Tx_j $. For Gaussian kernel compute $ k(x_i,x_j)= exp(\frac{-||x_i-x_j||_2} {2q^2})$.
 		\item [(iv)] Obtain the solution vectors $\alpha$, $\beta$  by solving the proposed QPP (\ref{dual_u_psvm1}).
 		\item [(v)] Also obtain the value of bias $b$ using (\ref{bb}).
 		\item [(vi)] Predict the label of test point $x$ using (\ref{pred}).
 	\end{algorithmic}
 \end{algorithm}
The proposed unified QPP (\ref{dual_u_psvm1}) should be solved for minimizing the pinball loss function in SVM for $ -1 \leq \tau \leq 1$.
 Some properties of Pin-SVM models like noise-insensitivity and non-sparsity have been only induced by the use of pinball loss function in the SVM model. Therefore, these properties do not vary in the proposed Unified Pin-SVM model. For clarity, we explicitly describe the algorithm of the proposed Unified Pin-SVM model in Algorithm 1. 
\section{Experimental Results}\label{sec4}        
In this section, we justify our claims made in this paper empirically. For this, we perform numerical experiments with some commonly real-world benchmark datasets. Table \ref{table1} shows the description of the used datasets in our experiments. The first four datasets in Table \ref{table1} contain the training and testing set provided. For other datasets, we have divided the training and testing set in Table \ref{table1}. We have normalized the training and testing set in $[-1,1]$.  
\begin{table}[]
		\caption{Dataset Description}
	\centering{\scriptsize
	\begin{tabular}{|l|l|l|}\hline
		Dataset & Size &  Training points  \\\hline 
		Monk 1   &  556$\times$ 7    & 124                       \\
		Monk 2	&  601 $\times$7     &  169                    \\
		Monk 3 & 554 $\times$ 7    &     122                   \\
		Spect &    267 $\times$ 22    &    80                      \\ \hline
		Fertility D. & 100 $\times$ 10  &   50                \\
		Echocardiogram & 131 $\times$ 10 &    80                \\ 
		Plrx & 182 $\times$ 13 & 100 \\  
		Sonar & 208 $\times$ 61  & 100 \\   
		Heart Statlog & 270 $\times$ 14   &    150                    \\
		Haberman &  306 $\times$ 4  &   150                    \\  
		Votes & 435 $\times$ 17 &    200               \\ 
		Ecoil  & 327 $\times$ 8 & 200 \\ 
		Ionosphere &  351 $\times$ 34     &  200               \\
		Bupa Liver &   345 $\times$ 7 &    250                \\
		Pima Indian &  768 $\times$ 9      &  300                \\
		Breast Cancer  & 569 $\times$ 31   &  400                          \\
		Australian  & 690  $\times$ 15   &  400                       \\ 
		Diabetes & 768 $\times$ 9 & 500 \\ 
		Spambase &  4601 $\times$ 57 & 4000 \\ \hline
	\end{tabular}}
	\label{table1}
\end{table}

Now, we describe our experimental setup. We have performed all experiments in MATLAB 2018 (\url{in.mathworks.com}) environment on a Dell Xeon processor with 16 GB of RAM and Windows 10 operating system. We have solved the dual QPP (\ref{dualpinsvm})  of the Pin-SVM model and the proposed dual  QPP (\ref{dual_u_psvm1}) of Unified Pin-SVM model with 'quadprog' function available in MATLAB. We have used linear kernel and RBF kernel of the form $exp(\frac{-||x-y||_2} {2q^2})$ in these QPPs of Pin-SVM models. The MATLAB codes of the proposed Unified Pin-SVM model and existing Pin-SVM models are available at
 \url{}{https://github.com/ltpritamanand/UnifiedPinSVM/\\tree/mycode/Unfied-Pin-SVM-master}.

Before reporting final numerical results, we have obtained the best possible choices of parameter $C_0$ and RBF kernel parameter $q$ of Pin-SVM models. For this, we have set $\tau=0$ in Pin-SVM model and searched best possible values of  $(C,q)$ from the set $\{ 2^ {-7},2^{-6},......,2^6,2^7\} \times \{ 2^ {-7},2^{-6},......,2^6,2^7\} $. After tunning the value of these parameters, we have obtained the accuracy of the Pin-SVM model and proposed Unified Pin-SVM model for different values of $\tau$ on different datasets listed in Table \ref{table1}.

\begin{figure*}
	\centering
	\subfloat[][Monk 1]{\includegraphics[width=5.0cm, height= 3.4cm]{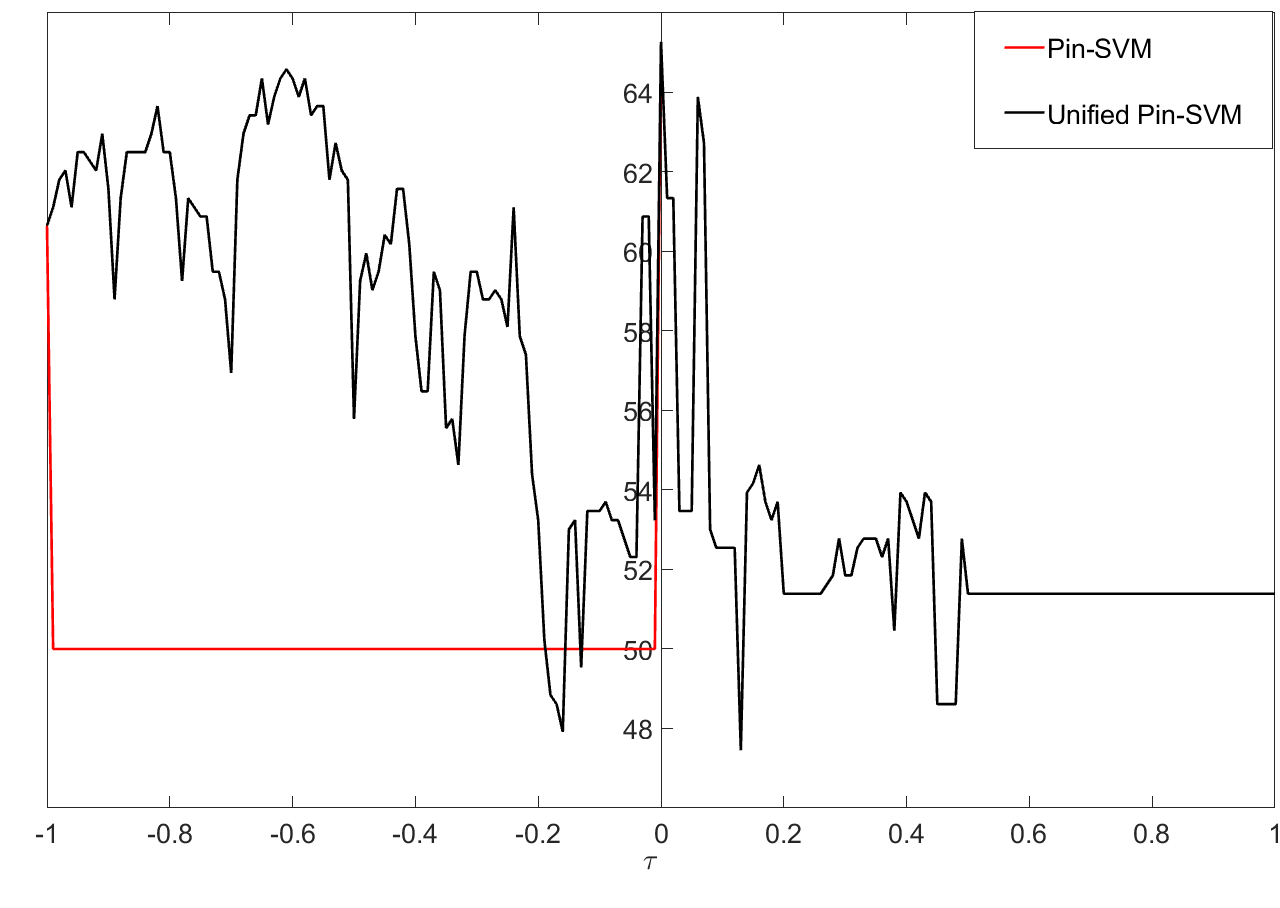}\label{<figure1>}}
    \subfloat[][Monk 3]{\includegraphics[width=5.0cm, height= 3.4cm]{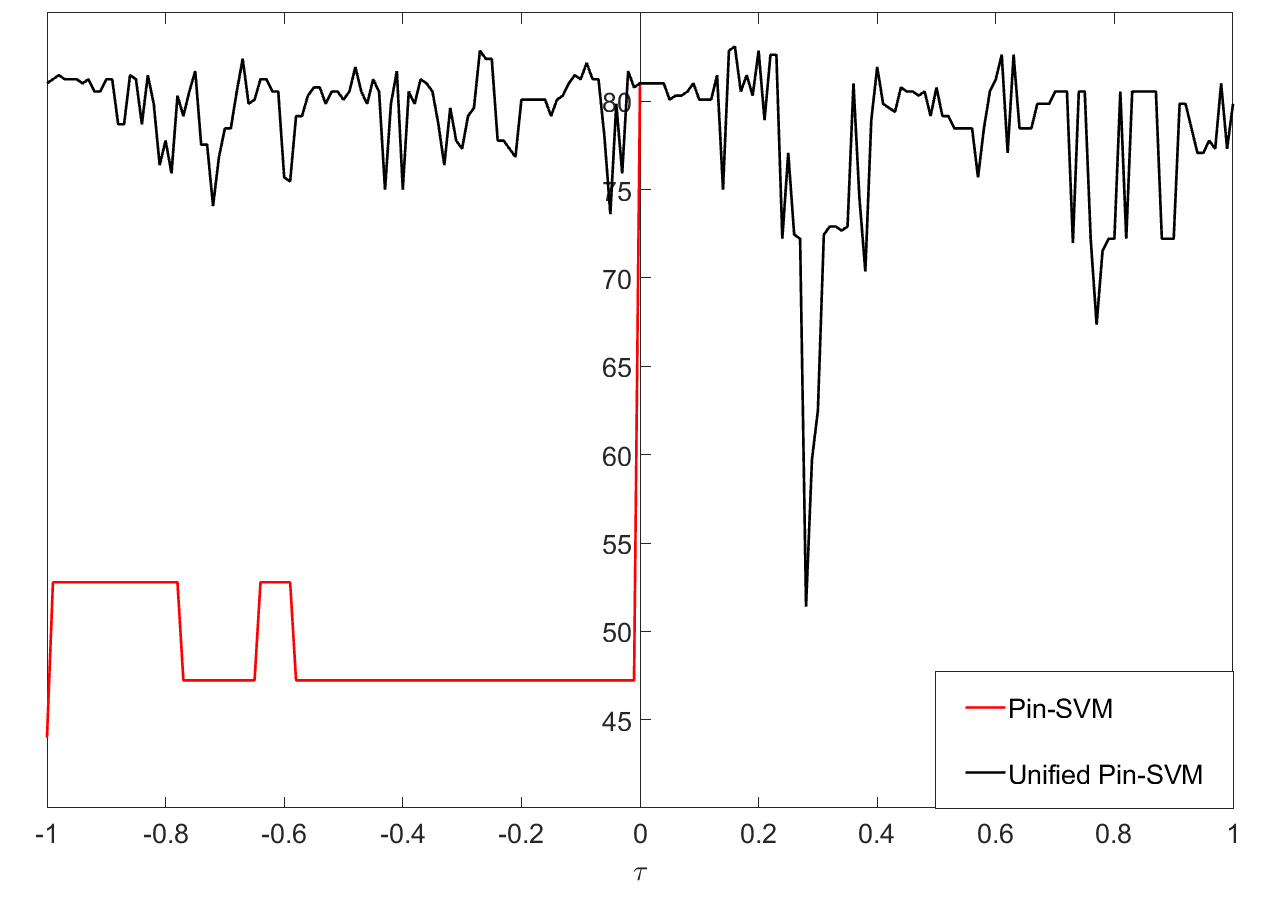}\label{<figure1>}}
	\subfloat[][Spect]{\includegraphics[width=5.0cm, height= 3.4cm]{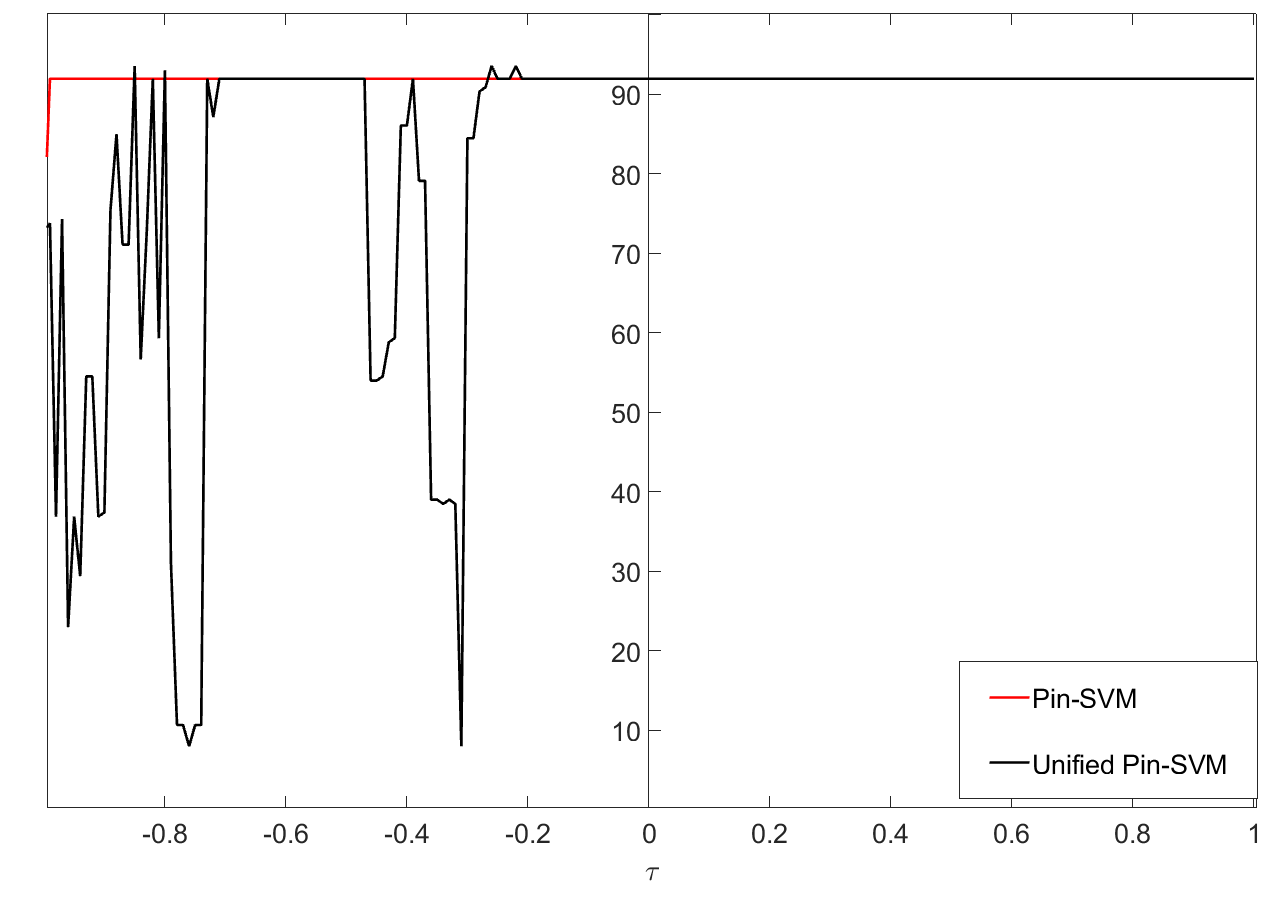}\label{<figure2>}}\\
	\subfloat[][Echo]{\includegraphics[width=5.0cm, height= 3.4cm]{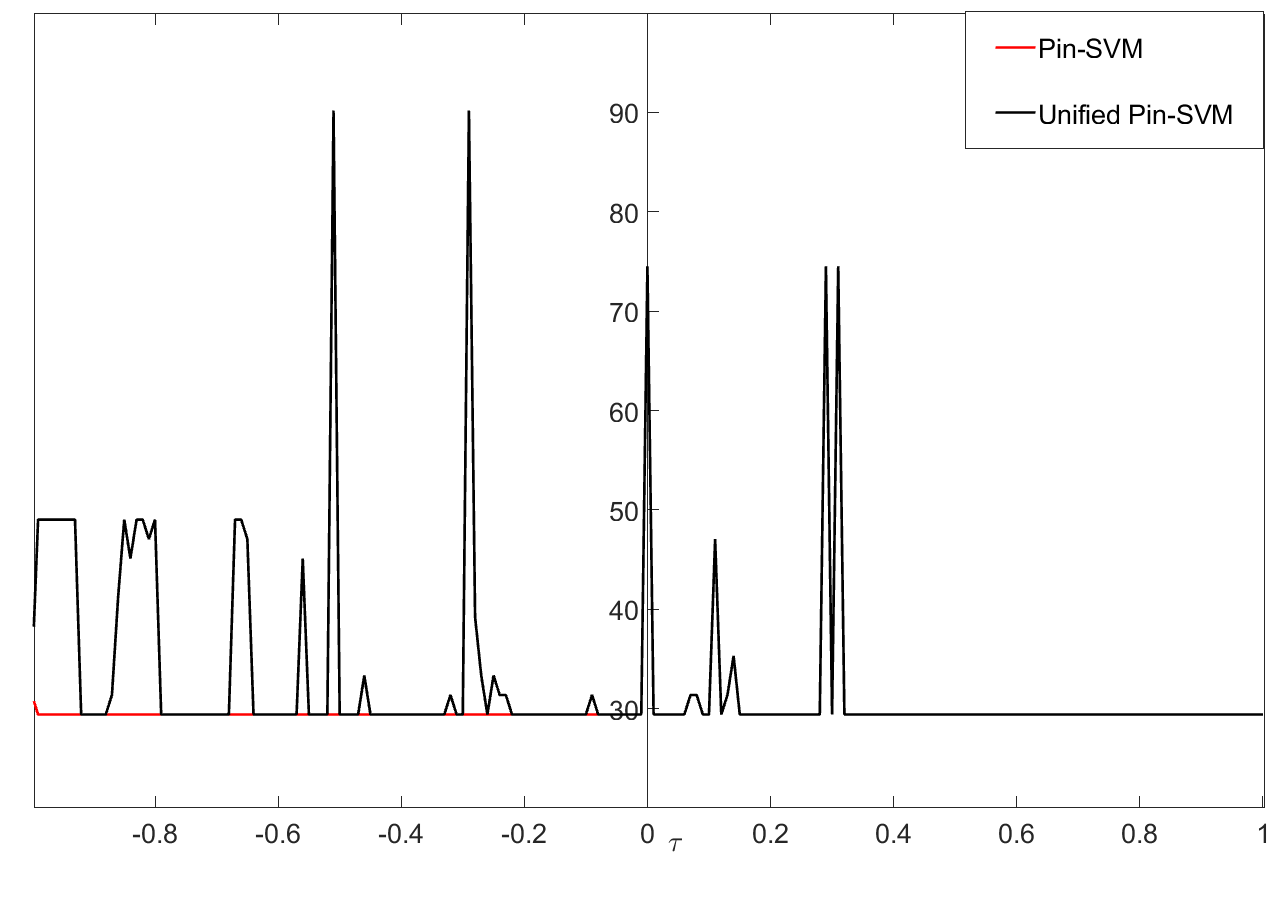}\label{<figure1>}}
	\subfloat[][Parlx]{\includegraphics[width=5.0cm, height= 3.4cm]{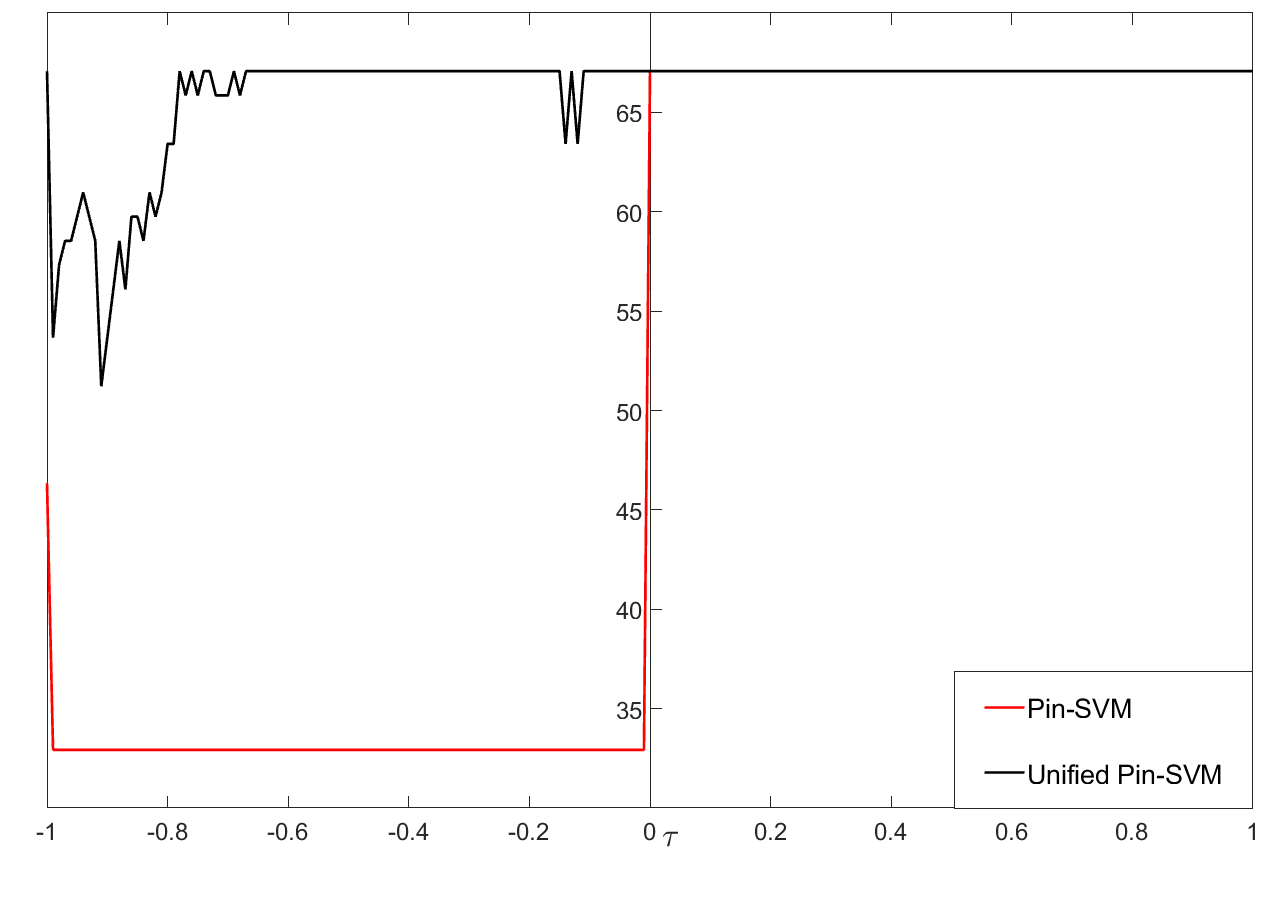}\label{<figure2>}}
	\subfloat[][Sonar]{\includegraphics[width=5.0cm, height= 3.4cm]{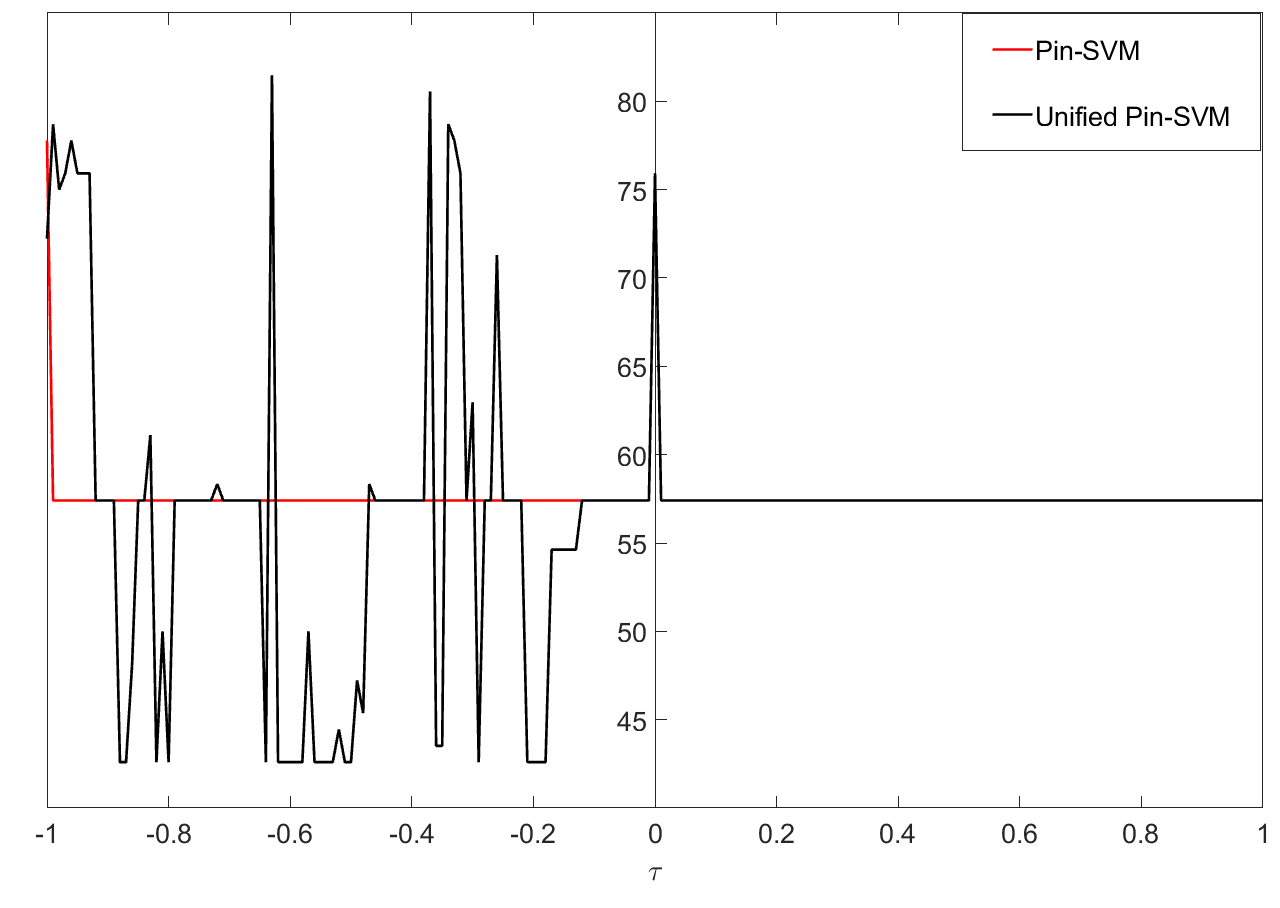}\label{<figure1>}}\\
	\subfloat[][Statlog]{\includegraphics[width=5.0cm, height= 3.4cm]{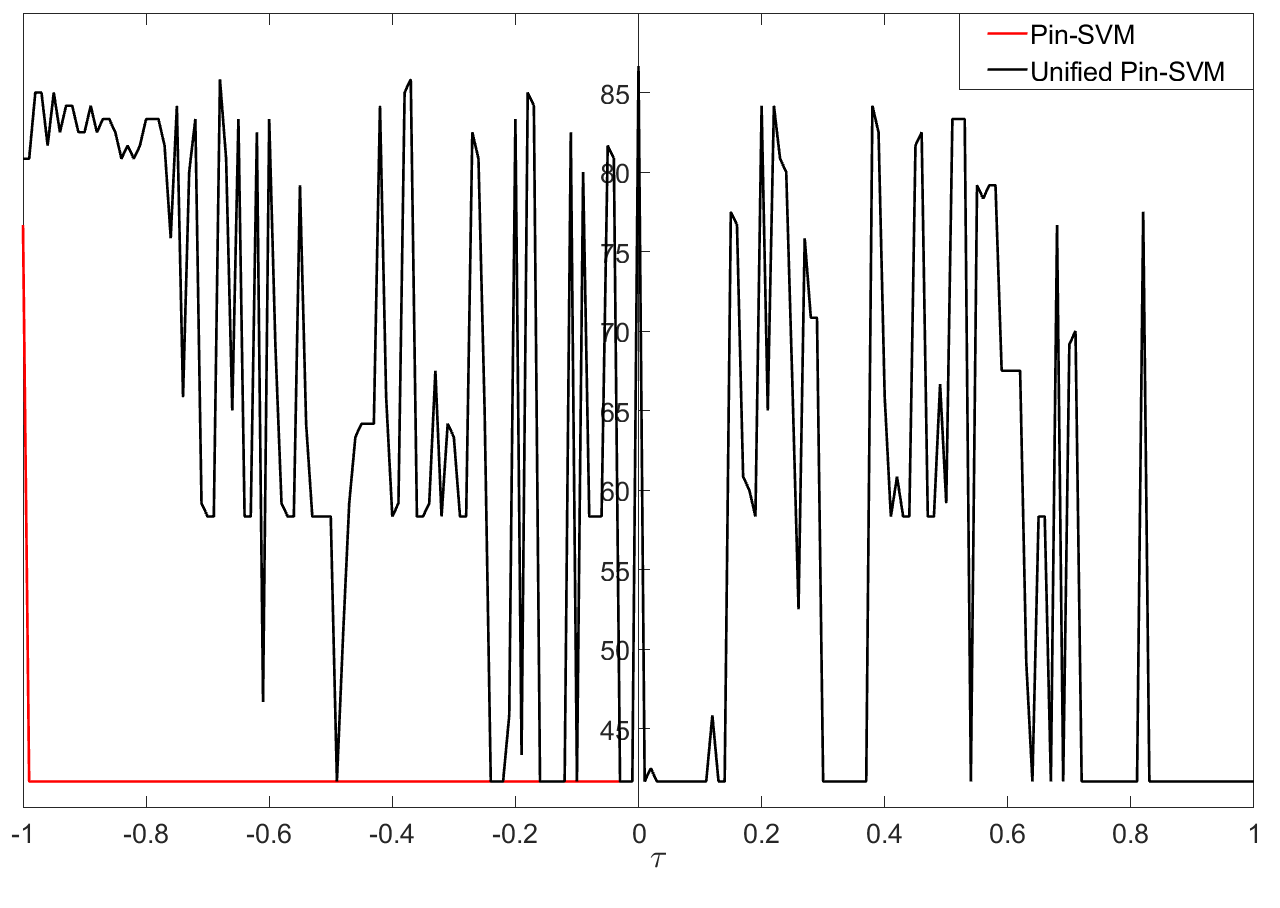}\label{<figure2>}}	
	\subfloat[][Ecoil]{\includegraphics[width=5.0cm, height= 3.4cm]{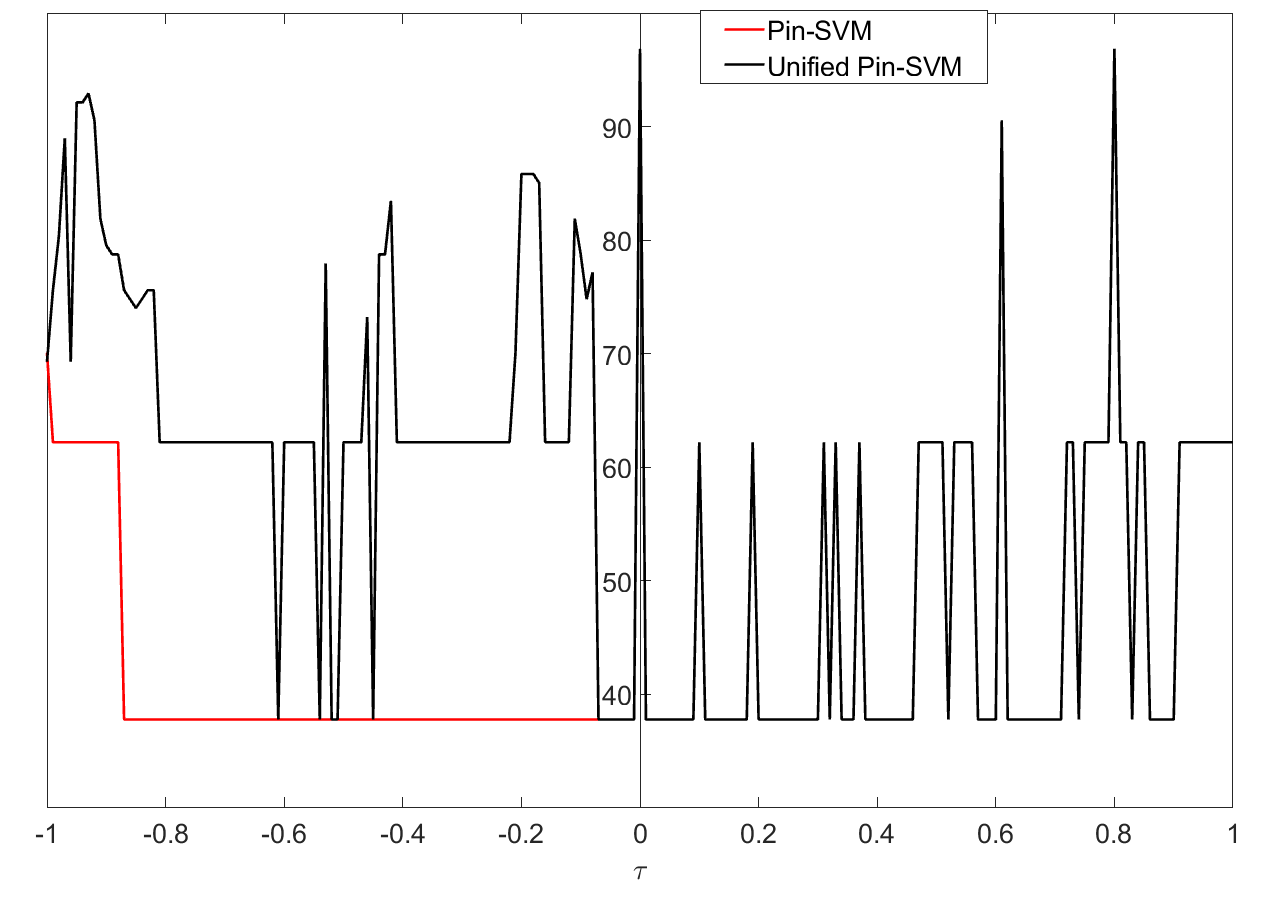}\label{<figure2>}}
	\subfloat[Bupa liver]{\includegraphics[width=5.0cm, height= 3.4cm]{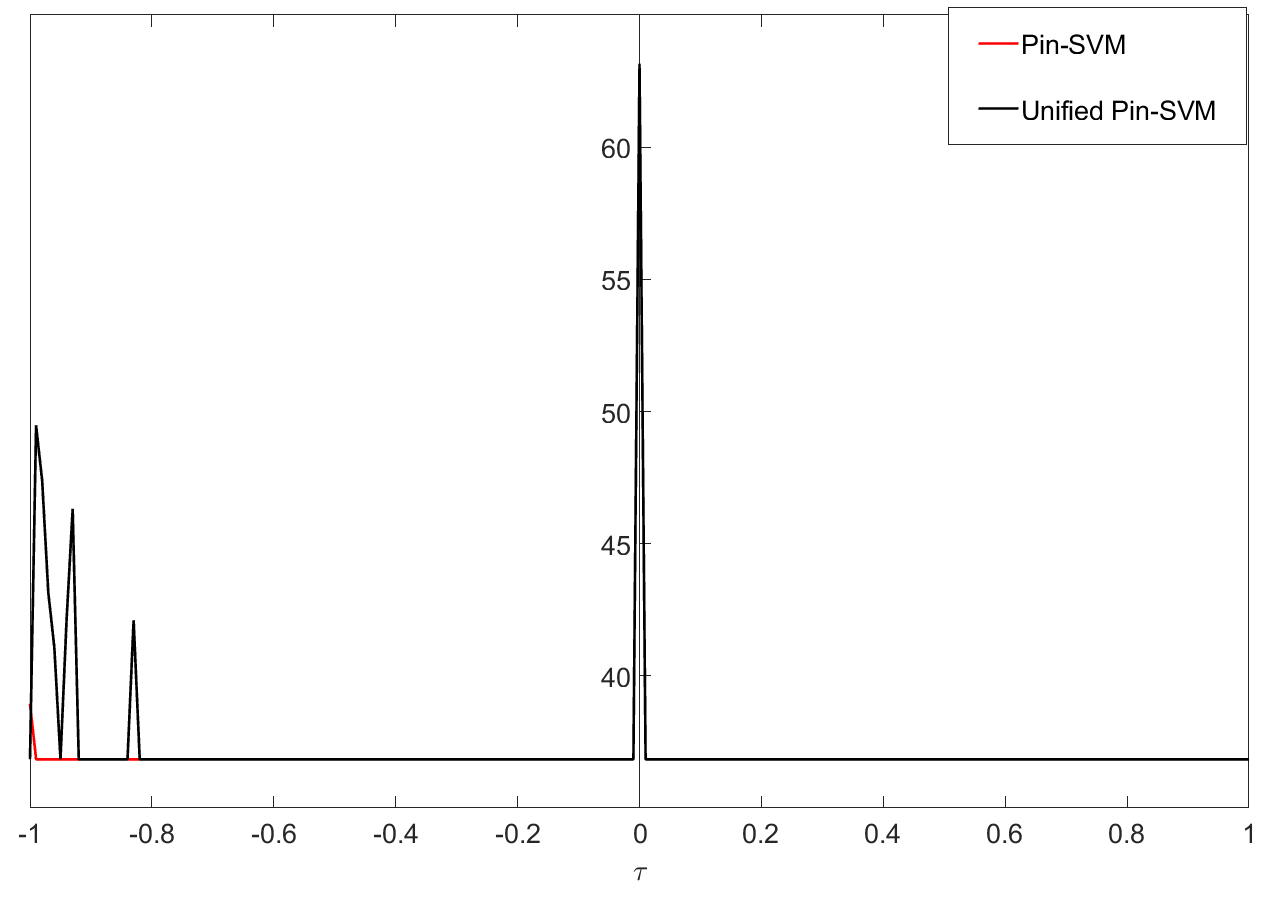}\label{<figure2>}}\\
	\subfloat[][Pima]{\includegraphics[width=5.0cm, height= 3.4cm]{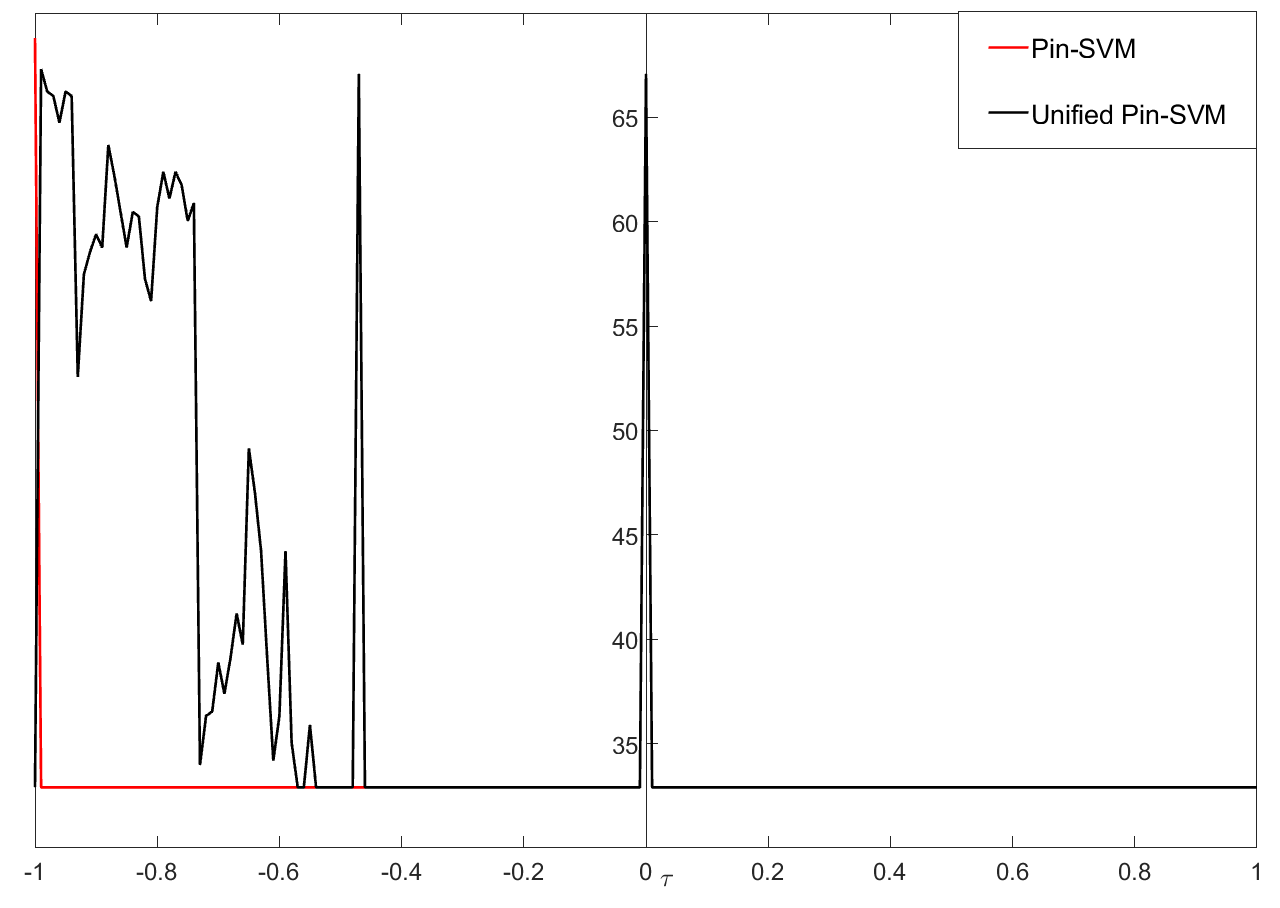}\label{<figure1>}}
	\subfloat[][Breast Cancer]{\includegraphics[width=5.0cm, height= 3.4cm]{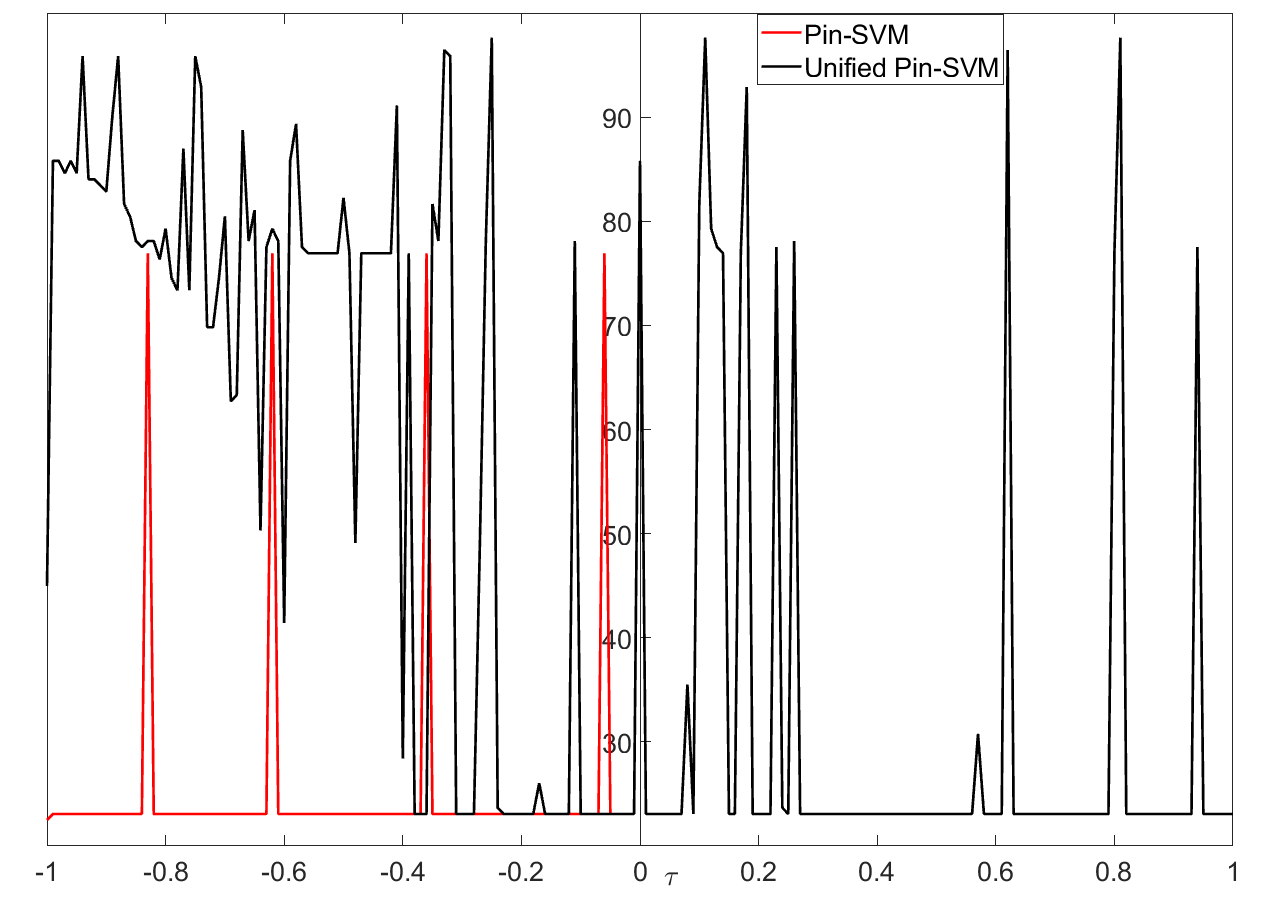}\label{<figure2>}}
	\subfloat[][Australian]{\includegraphics[width=5.0cm, height= 3.4cm]{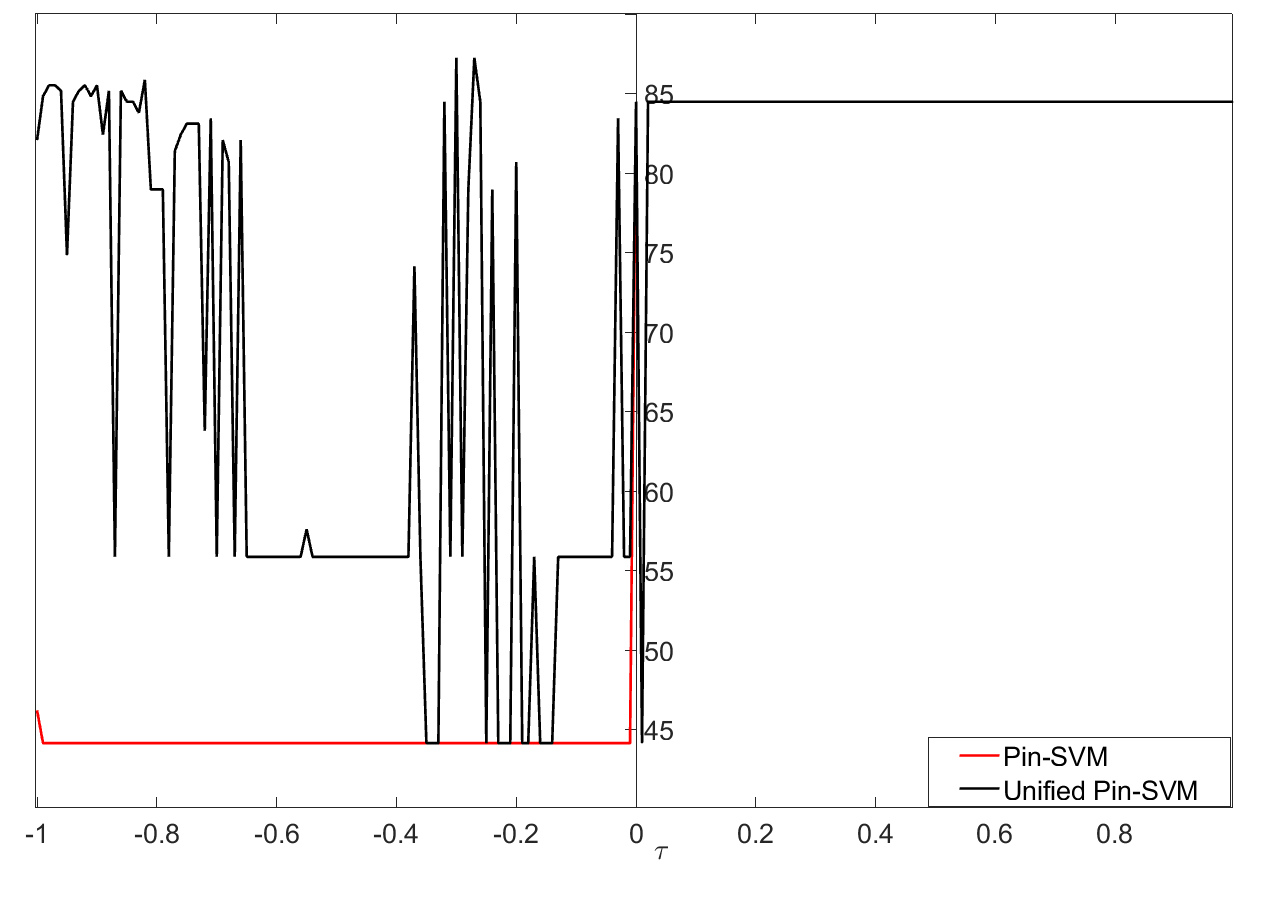}\label{<figure1>}}\\
	\subfloat[][Diabetes]{\includegraphics[width=5.0cm, height= 3.4cm]{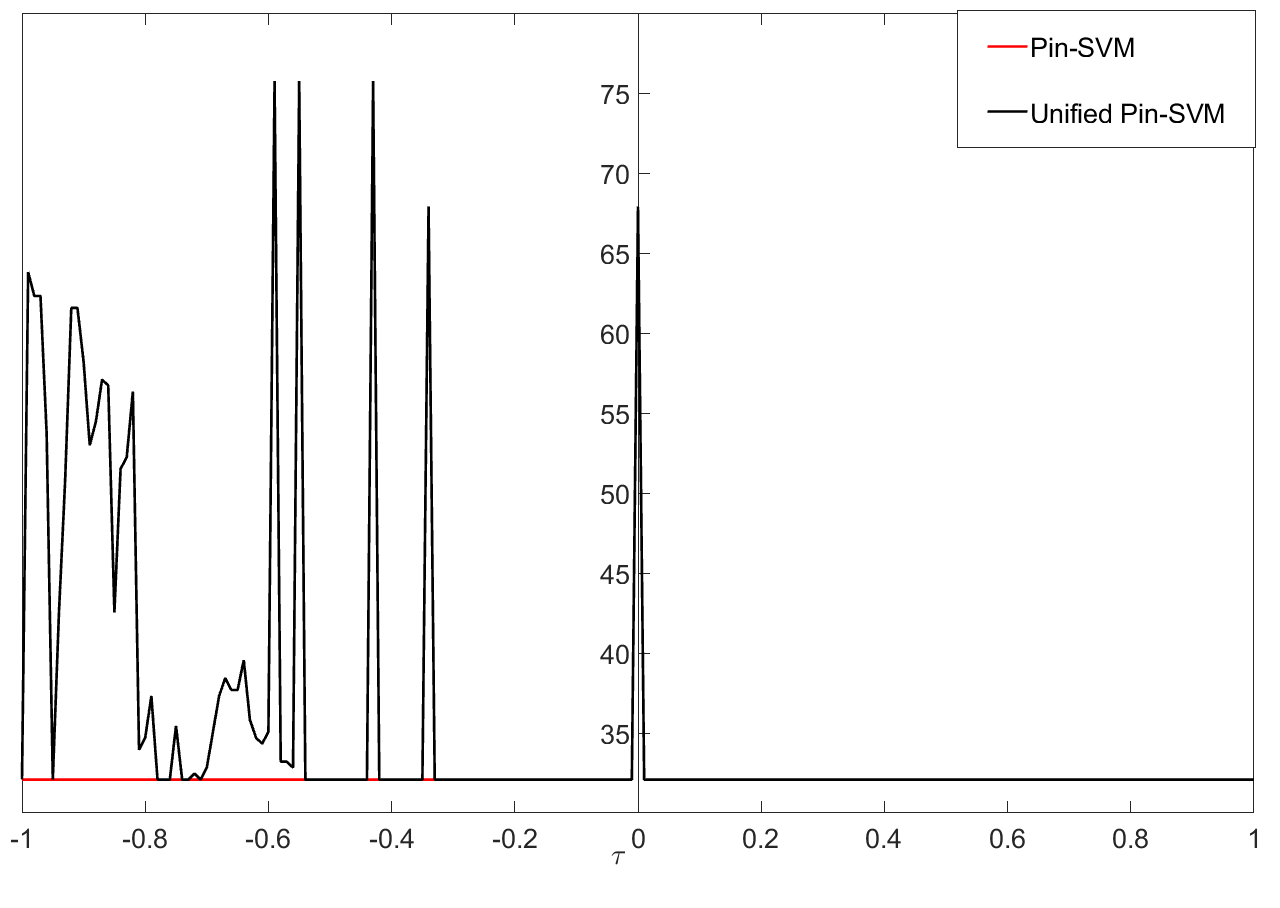}\label{<figure2>}} 
\subfloat[][Spambase]{\includegraphics[width=5.0cm, height= 3.4cm]{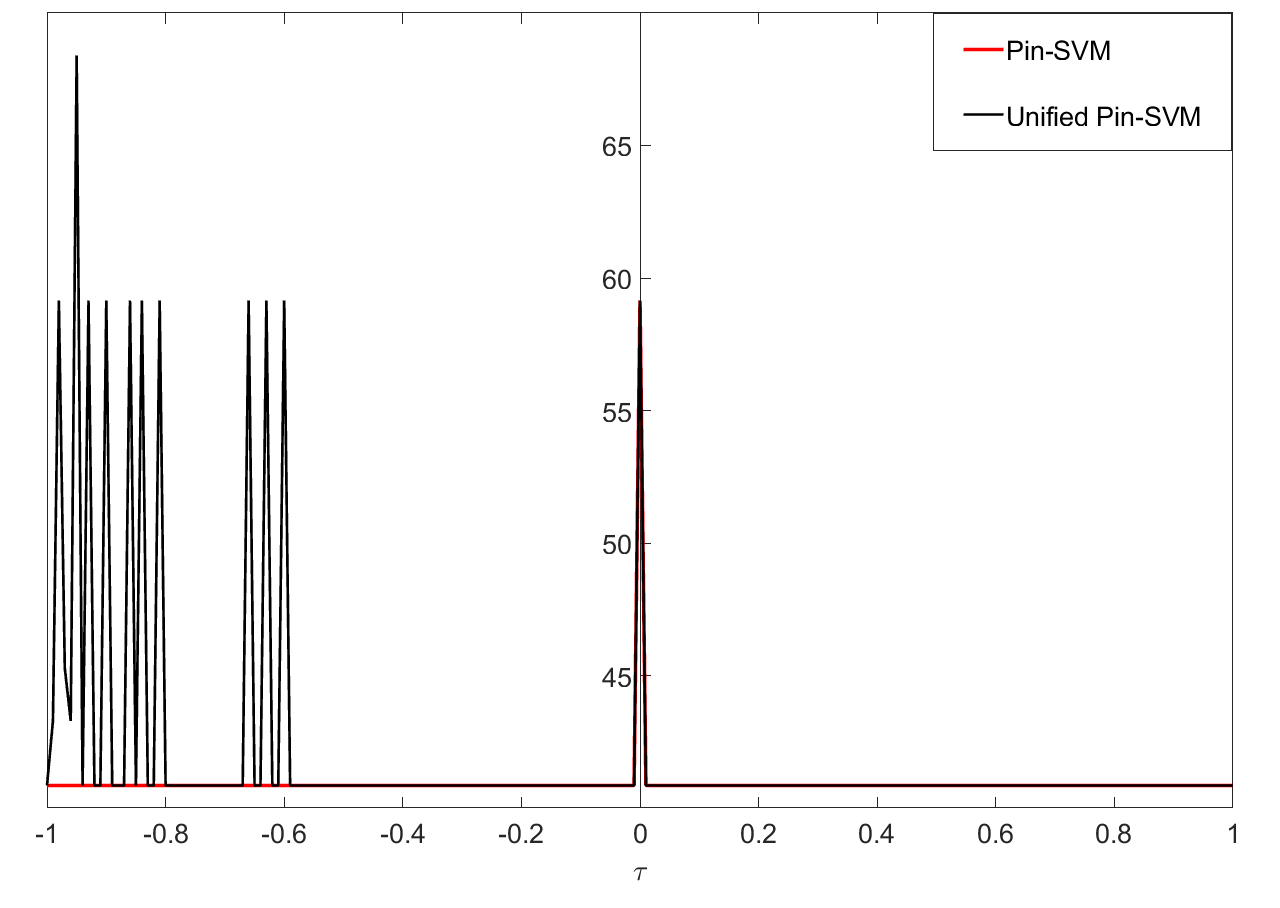}\label{<figure1>}}
\subfloat[][Fertility]{\includegraphics[width=5.0cm, height= 3.4cm]{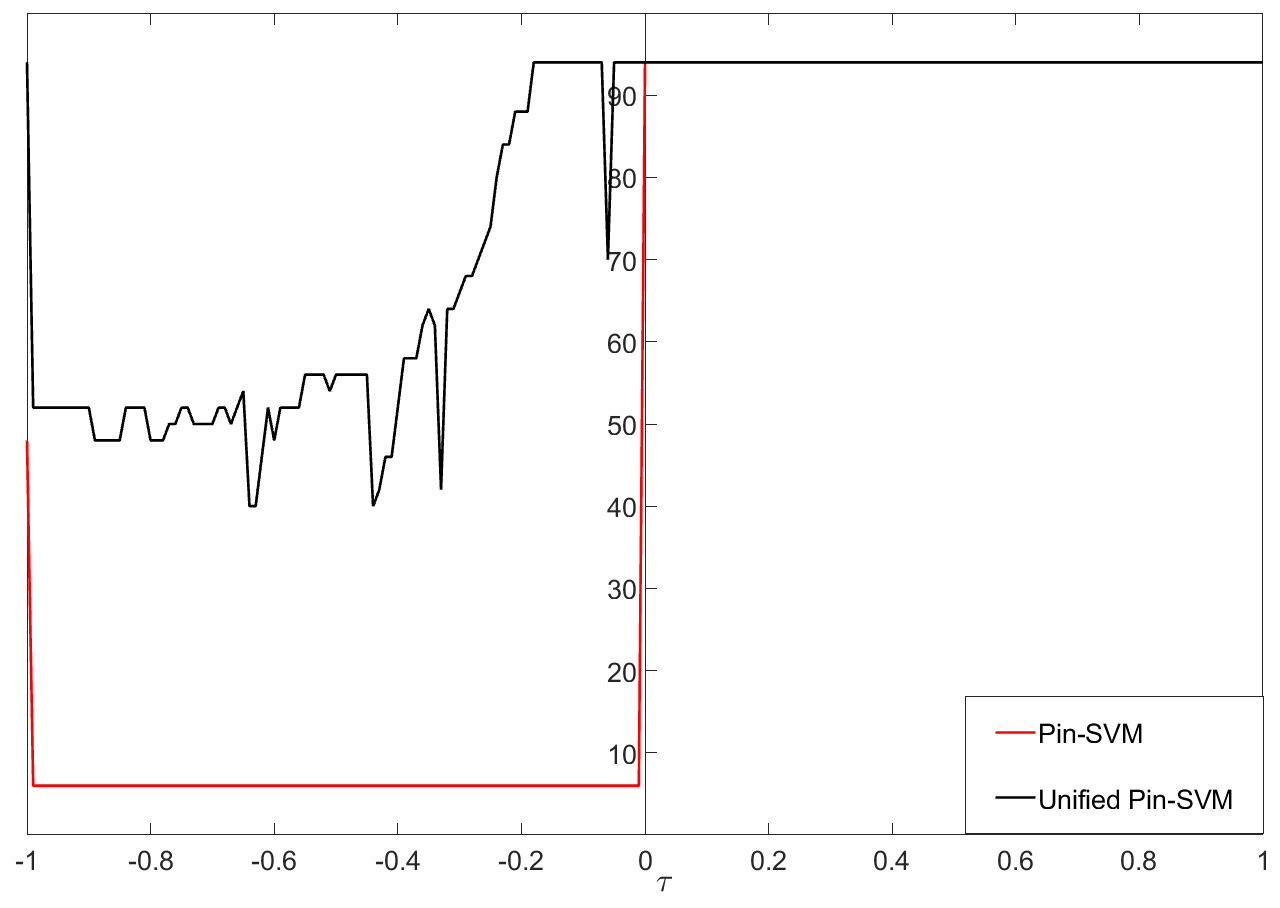}\label{<figure1>}}
	\caption{Comparison of existing Pin-SVM model and proposed Unified Pin-SVM model with linear kernel.}
	\label{steady_state1}
\end{figure*}

Figure \ref{steady_state1} shows the plot of  accuracy on several datasets obtained by the existing Pin-SVM model and proposed Unified Pin-SVM model against different $\tau$ values from the set $\{ -1,-0.99,....,..0.99,1\}$ with linear kernel. In these plots, the red-line represents the accuracy obtained by Pin-SVM and the black line represents the accuracy obtained by the proposed Unified Pin-SVM model. It should be noted that at $\tau=0$, the Pin-SVM and proposed Unified Pin-SVM model reduces to the $C$-SVM model. We can obtain the following observations from plots in Figure \ref{steady_state1}. 

\begin{enumerate}
	\item  In each plot, we can observe that the black line hides the red line on the right side of the $Y$-axis. It confirms that for $\tau \geq 0$ , the Pin-SVM model and proposed Unified SVM model are equivalent. 
	\item In each plot, the red line differs from the black line in left side of the $Y$-axis. It empirically confirms that the Pin-SVM model (\ref{opt_for_pin_neg}) for $ -1 \leq \tau < 0$  is different from the Unified Pin-SVM model for $ -1 \leq \tau < 0$ .
	\item  Further, we can observe that the black line appears above the red line on the left side of $Y$-axis in most of the cases. It means that for $ -1 \leq \tau < 0$, the proposed  Unified Pin-SVM can obtain better accuracy than the existing Pin-SVM model (\ref{opt_for_pin_neg}). It is because of the fact that Unified Pin-SVM minimizes the pinball loss function for $ -1 \leq \tau < 0$  in true spirit.
\end{enumerate}
We have also plotted the accuracy obtained by the proposed Unified Pin-SVM model and existing Pin-SVM model against different values of parameter $C$ and $\tau$ in  Figure \ref{fig3d} for few datasets.  For this, we have varied $\tau$ and C in the range $\{ -1,-0.9,...,0.9,1\}$ and $\{2^{-7}, 2^{-6},...,2^{6},2^7\}$ respectively. Figure \ref{fig3d} confirms that irrespective of choice of parameters, the proposed Unified Pin-SVM model outperforms the existing Pin-SVM model for $ -1 \leq \tau < 0$.

\begin{figure*}[h]
	\centering
	\subfloat[][Monk 1]{\includegraphics[width=5.0cm, height= 3.4cm]{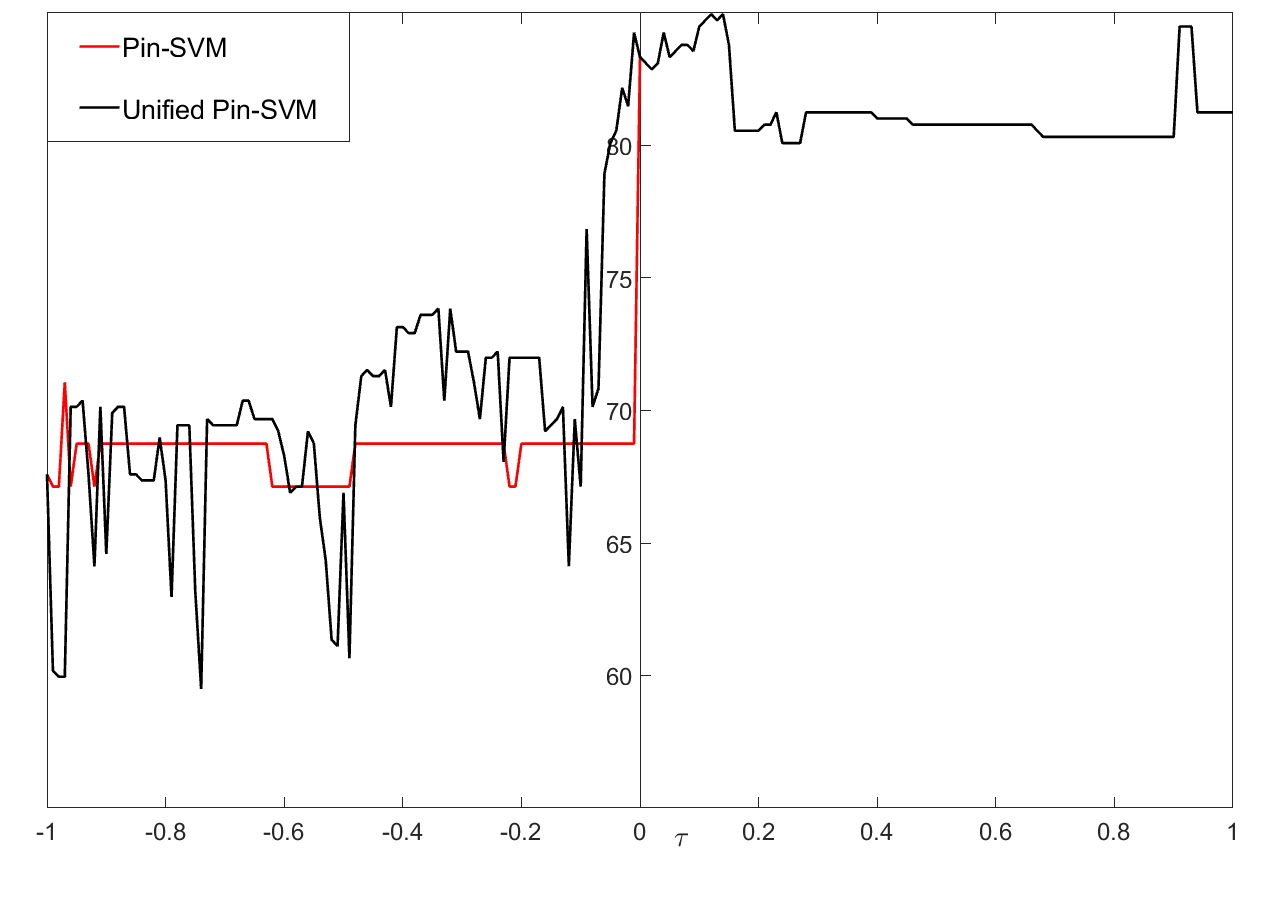}\label{<figure1>}}
	\subfloat[][Monk 2]{\includegraphics[width=5.0cm, height= 3.4cm]{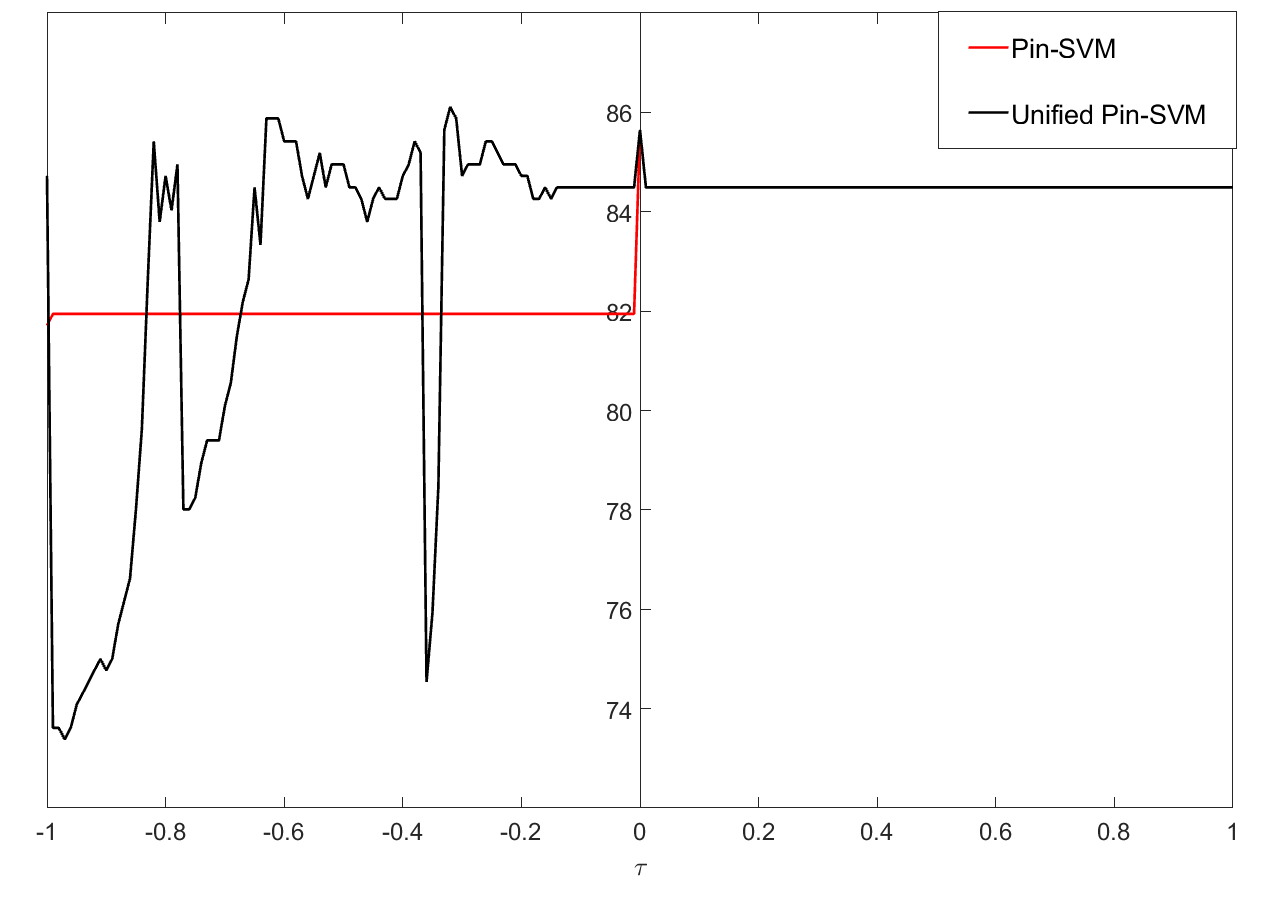}\label{<figure1>}}
	\subfloat[][Monk 3]{\includegraphics[width=5.0cm, height= 3.4cm]{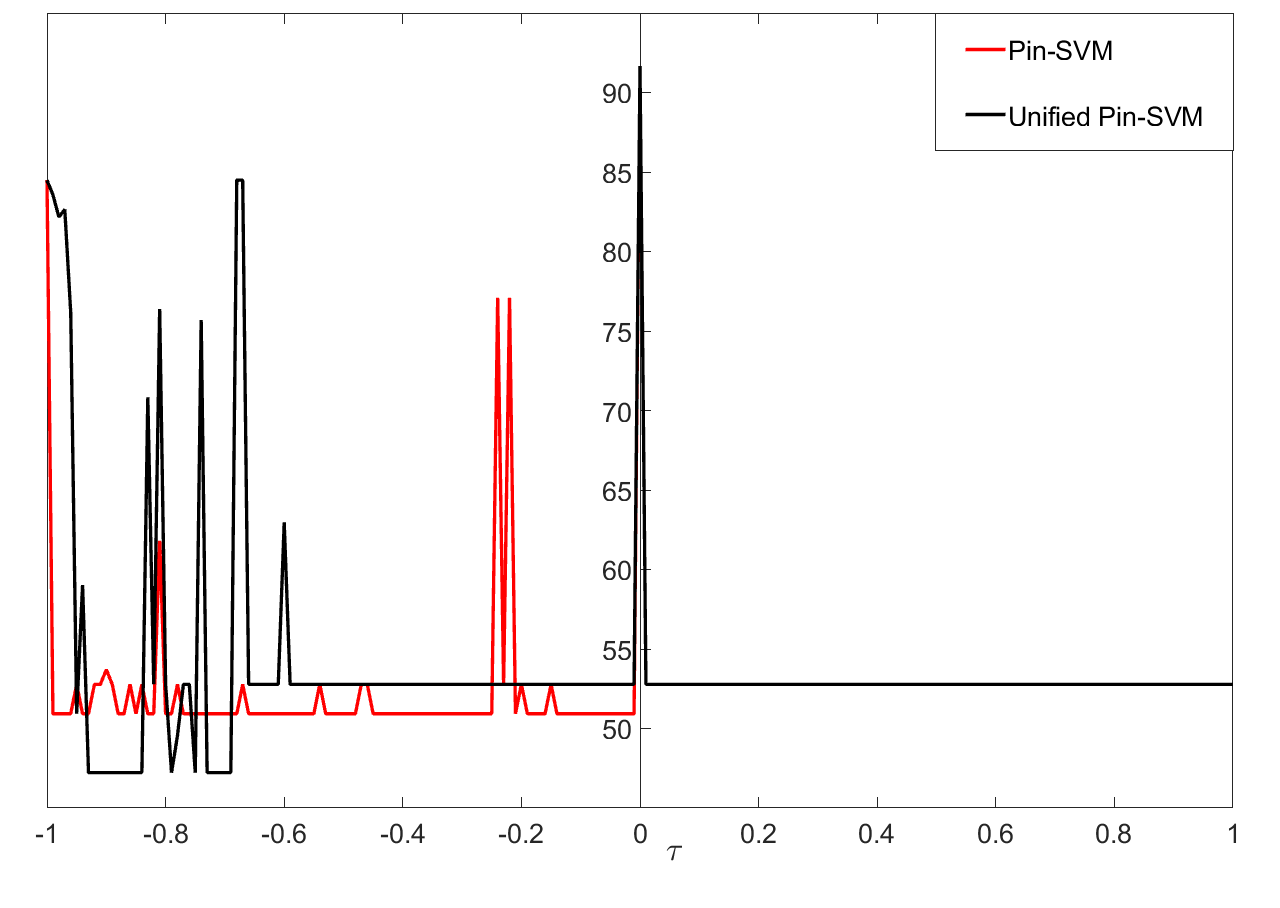}\label{<figure1>}}\\
	\subfloat[][Spect]{\includegraphics[width=5.0cm, height= 3.4cm]{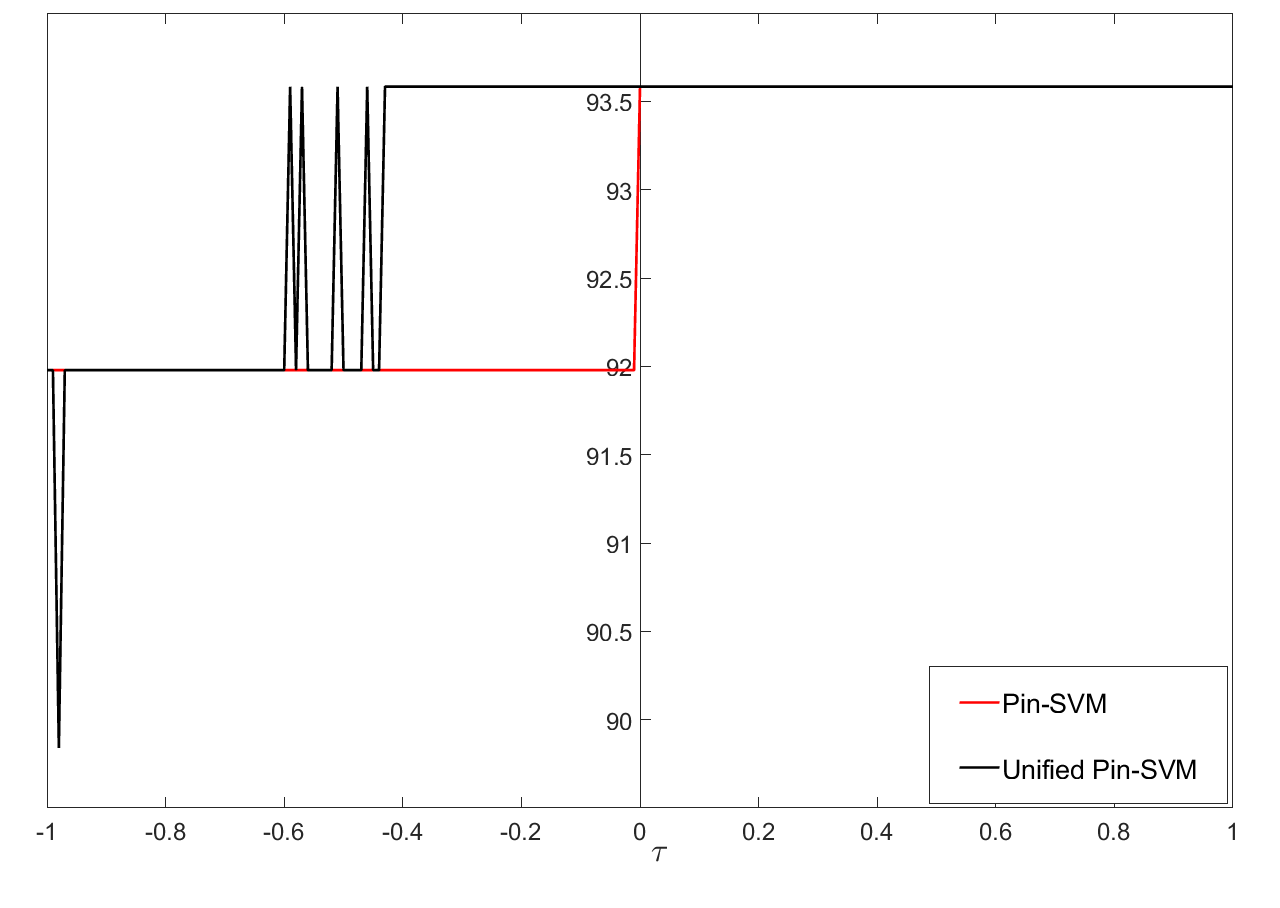}\label{<figure2>}}
	\subfloat[Bupa liver]{\includegraphics[width=5.0cm, height= 3.4cm]{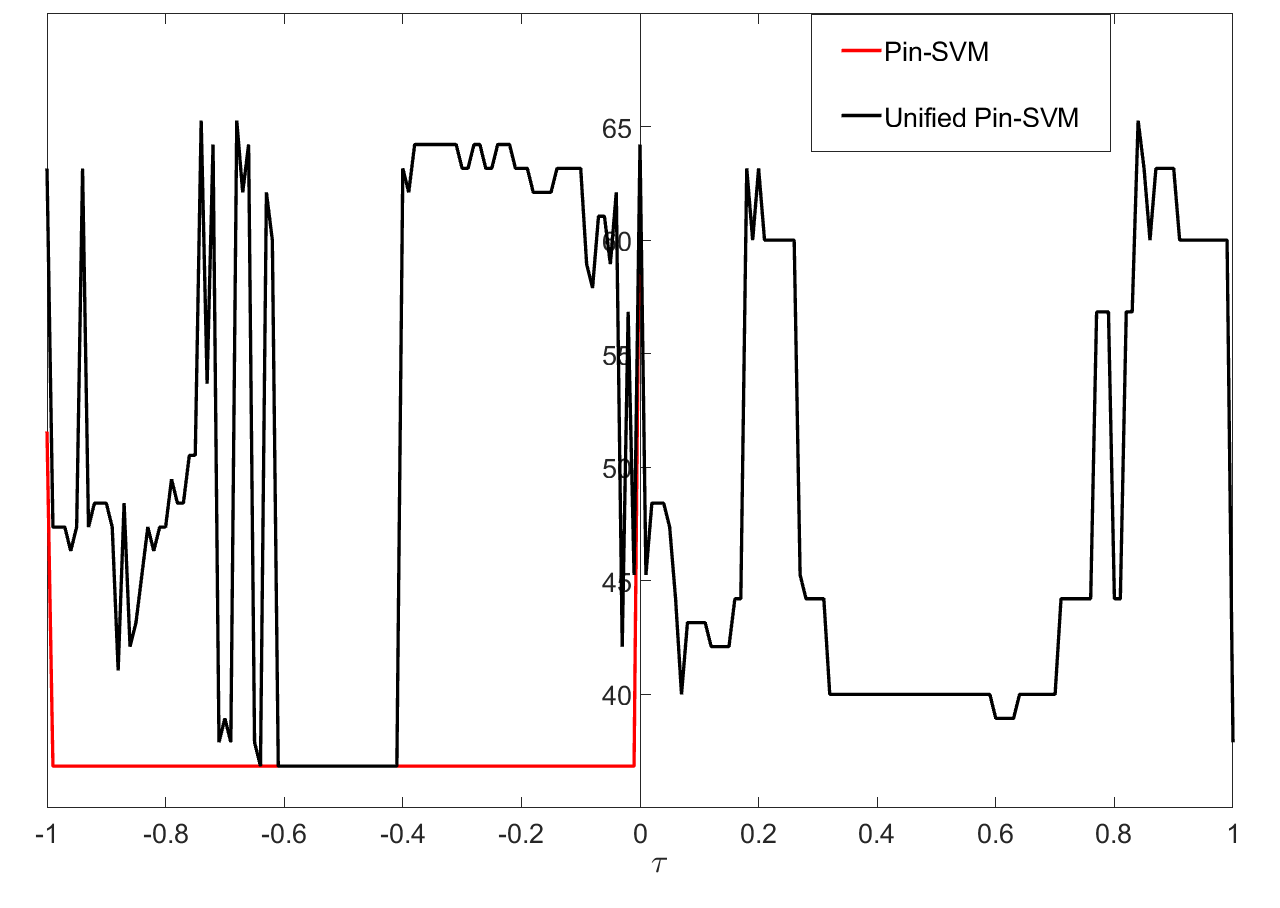}\label{<figure2>}}
	\subfloat[][Pima]{\includegraphics[width=5.0cm, height= 3.4cm]{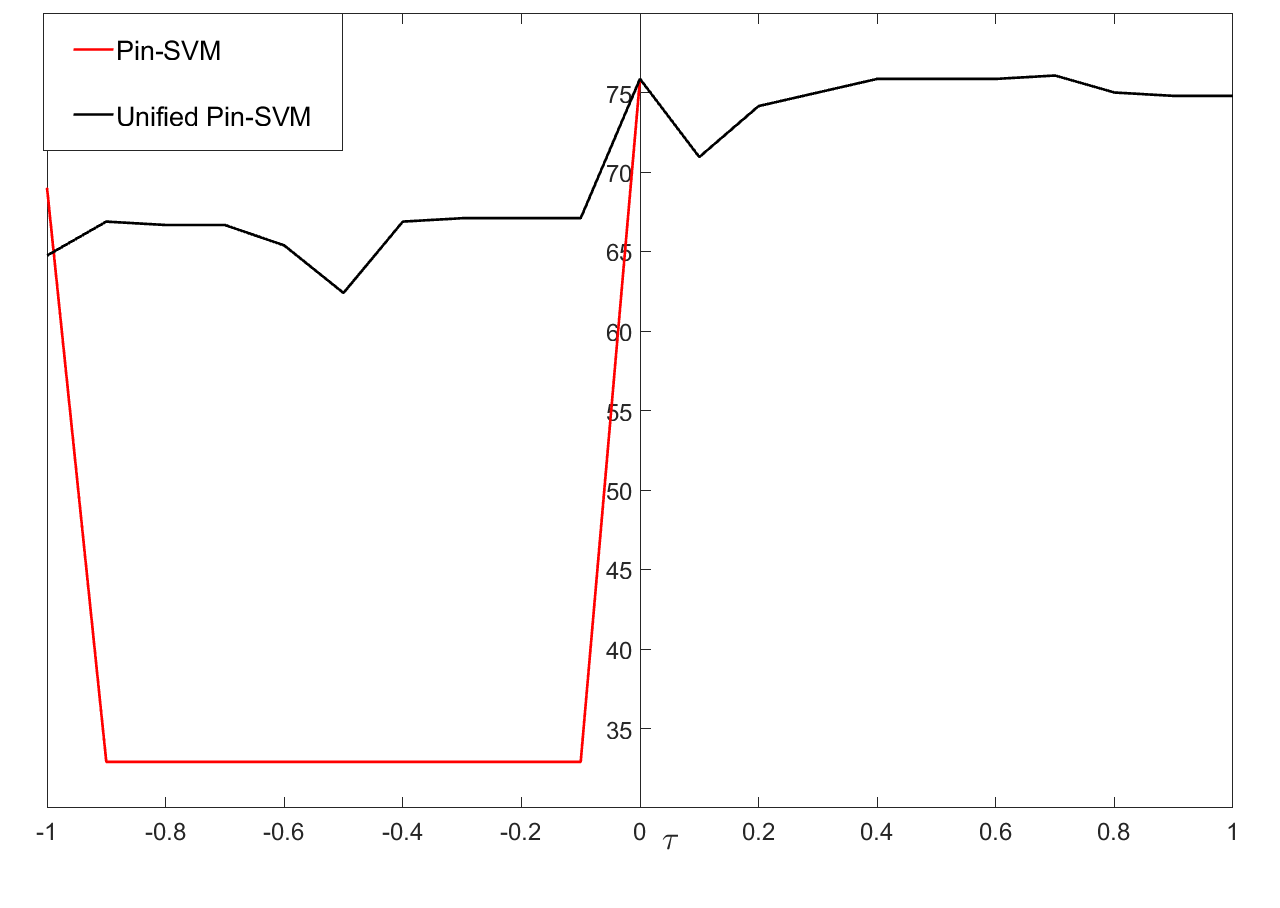}\label{<figure1>}}\\
	\subfloat[][German]{\includegraphics[width=5.0cm, height= 3.4cm]{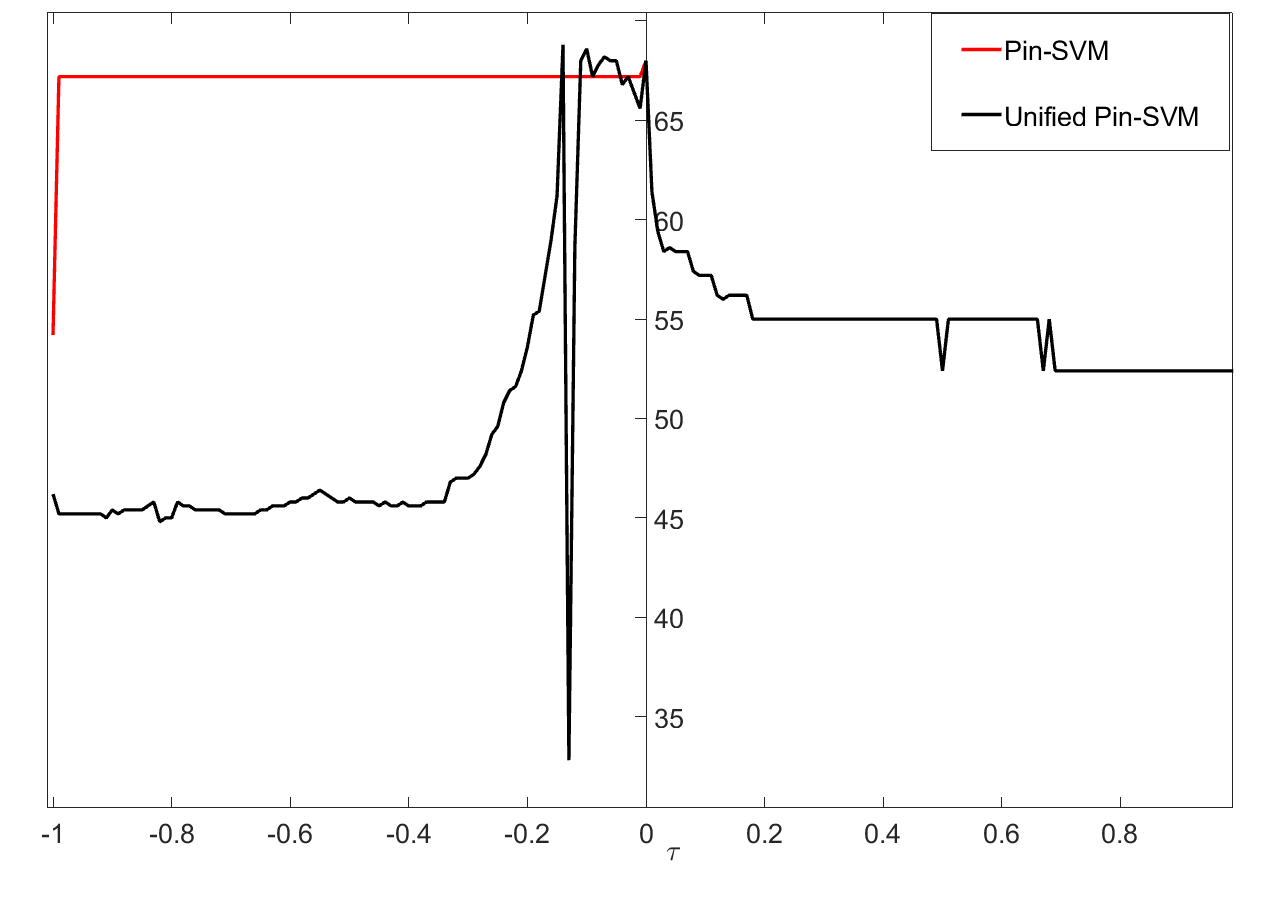}\label{<figure2>}}
	\subfloat[][Australian]{\includegraphics[width=5.0cm, height= 3.4cm]{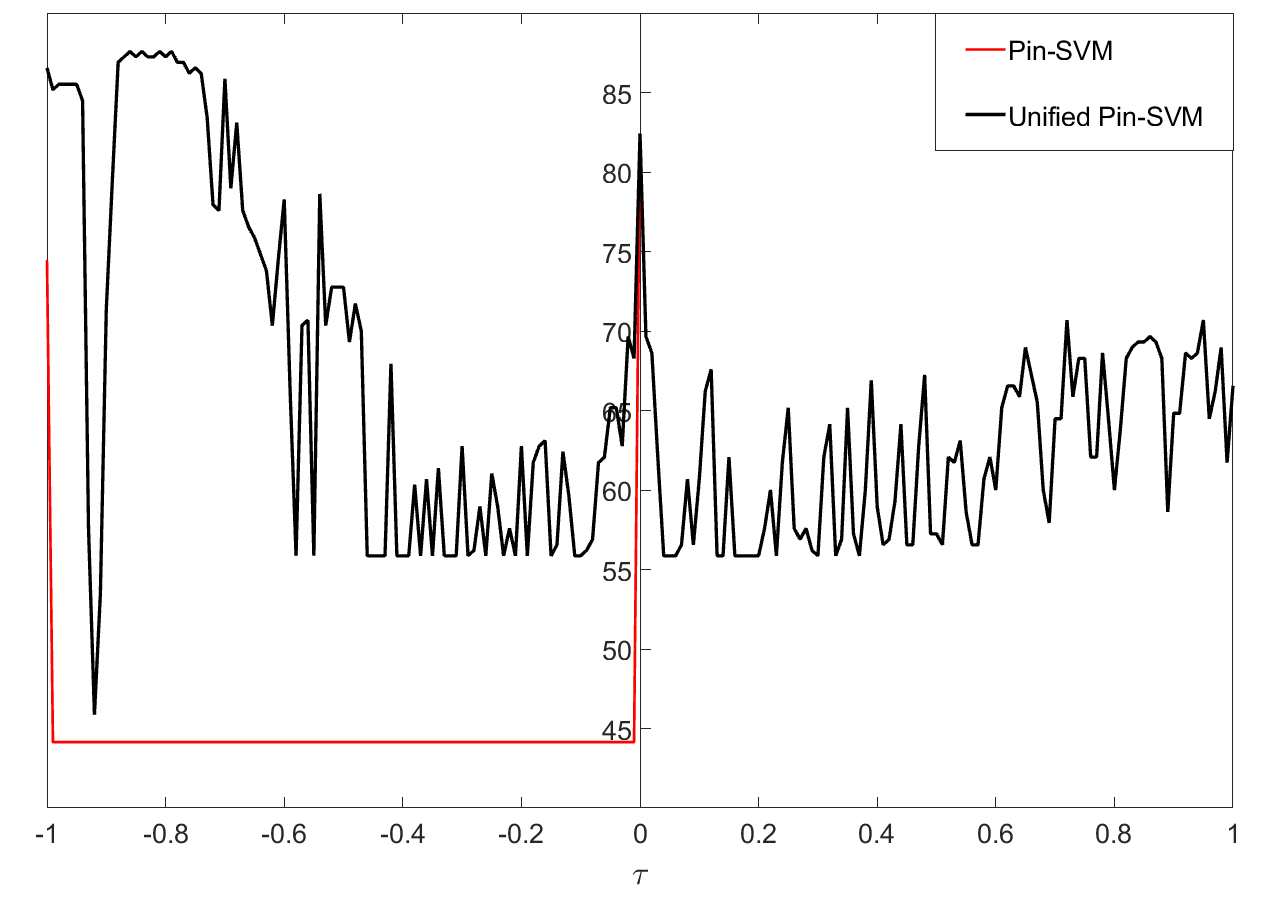}\label{<figure1>}}
	\subfloat[][Diabetes]{\includegraphics[width=5.0cm, height= 3.4cm]{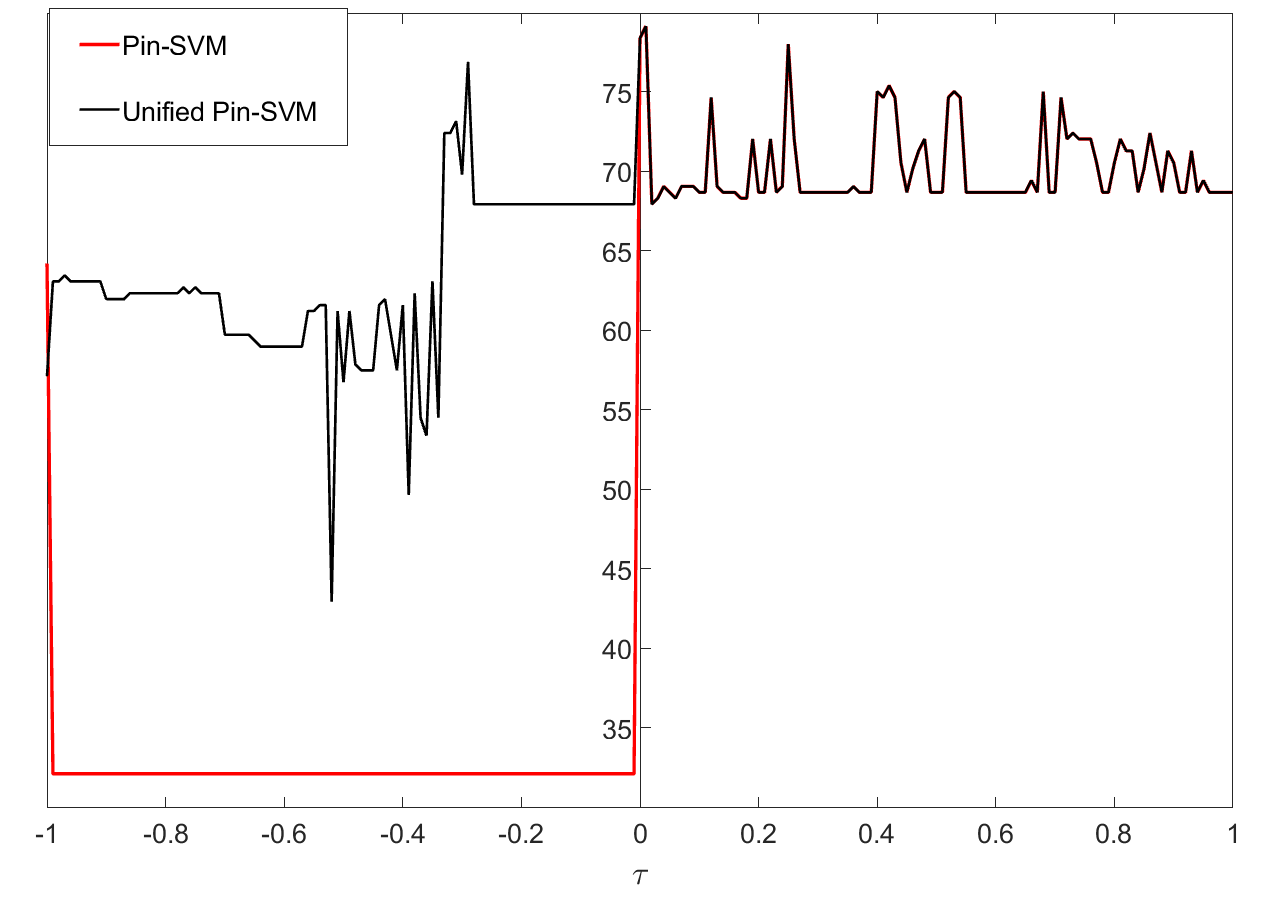}\label{<figure2>}} 
	\caption{Comparison of existing Pin-SVM model and proposed Unified Pin-SVM model with RBF kernel.}
	\label{steady_state}
\end{figure*}
\begin{figure*}
	\centering
	\subfloat[][Monk 1]{\includegraphics[width=6.5cm, height= 4.2cm]{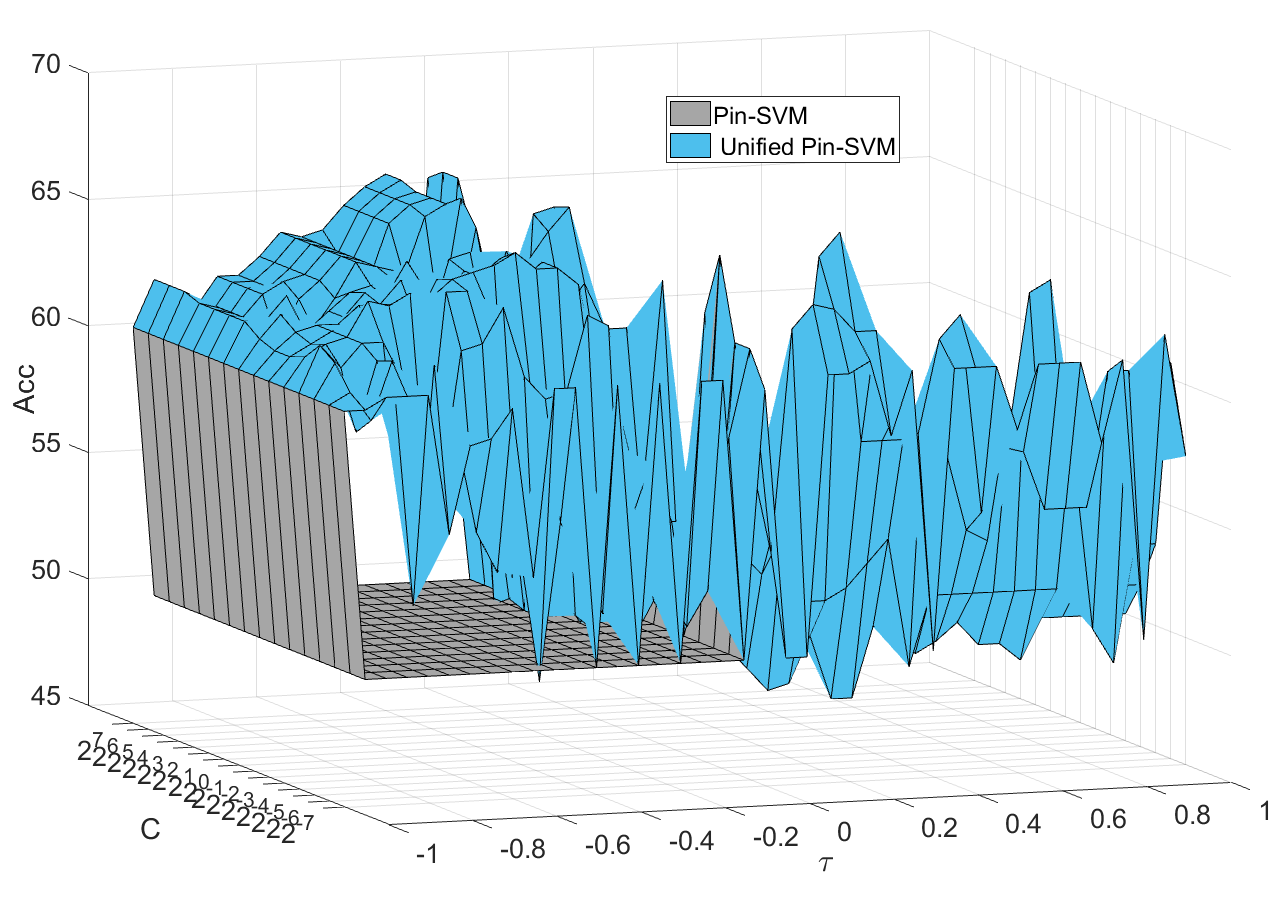}} 
	\subfloat[][Statlog]{\includegraphics[width=6.5cm, height= 4.2cm]{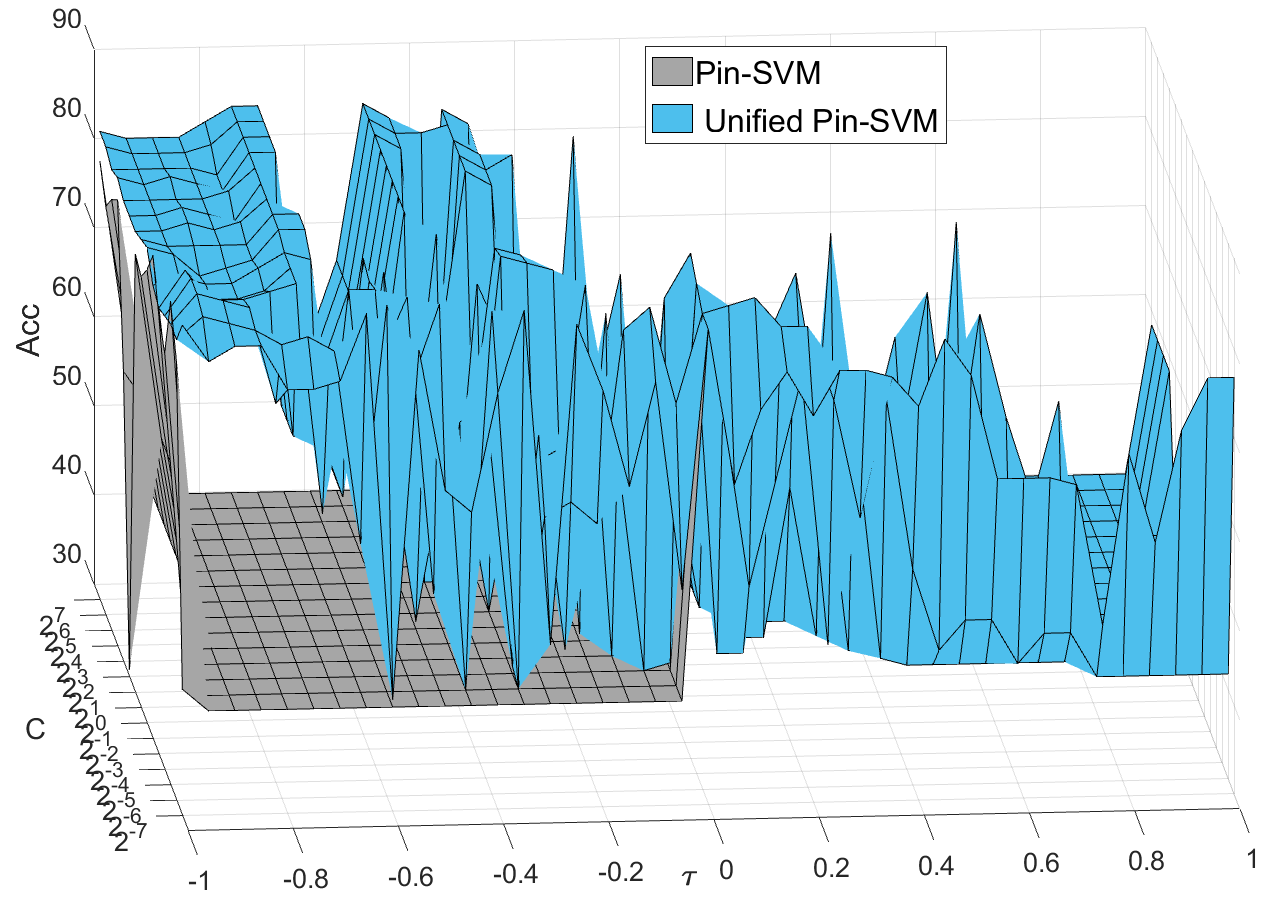}}
	\caption{ Plot of accuracy obtained by Pin-SVM models for different values of its parameters $\tau$ and  $C_o$. }
	\label{fig3d}
\end{figure*}

\begin{table}
\caption{Pin-SVM models with  linear kernel}
{\scriptsize 	\begin{tabular}{|c|c|c|c|c|c|}\hline
		
Dataset&     SVM models & 	Accuracy   & 	Time(s)  & $\tau$\\ \hline
Monk1      & Unified Pin-SVM & 65.28                          & 0.17                                       & 0.00                      \\
 $C_0= 0.0625$ & Pin-SVM         & 65.28                          & 0.16                                     & 0.00                      \\
& C-SVM             & 65.28                          & 0.15                                       & -                      \\
\hline
  Monk2      & Unified Pin-SVM & 67.13                          & 0.29                                   & -0.60                     \\
$C_0= 0.0078$ & Pin-SVM         & 67.13                          & 0.29                                     & -0.99                     \\
& SVM             & 67.13                          & 0.22                                     & -                      \\
\hline
Monk3      & Unified Pin-SVM & 83.10                          & 0.17                                    & 0.16                      \\
$C_0= 0.0078$ & Pin-SVM         & 83.10                          & 0.17                                      & 0.16                      \\
& C-SVM             & 81.02                          & 0.15                                       & -                     \\
\hline
Spect      & Unified Pin-SVM & 93.58                          & 0.07                                      & -0.85                     \\
 $C_0= 0.0156$& Pin-SVM         & 91.98                          & 0.08                                     & -0.99                     \\
&C-SVM             & 91.98                          & 0.05                                      & -                      \\
\hline
Haberman   & Unified Pin-SVM & 76.28                          & 0.19                          & -0.61                     \\
$C_0= 0.0078$& Pin-SVM         & 73.08                          & 0.11                                  & 0.00                      \\
& C-SVM             & 73.08                          & 0.10                                      & -                      \\
\hline
Heart Statlog     & Unified Pin-SVM & 86.67                          & 0.09                                     & 0.00                      \\
$C_0= 0.0625$& Pin-SVM         & 86.67                          & 0.09                                     & 0.00                      \\
&C-SVM             & 86.67                          & 0.09                                       & -                     \\
\hline
Ionosphere       & Unified Pin-SVM & 94.04                          & 0.16                                     & 0.00                      \\
$C_0= 2$& Pin-SVM         & 94.04                          & 0.16                                     & 0.00                      \\
& C-SVM             & 94.04                          & 0.16                                       & -                    \\
\hline
Pima       & Unified Pin-SVM & 67.31                          & 0.89                                       & -0.99                     \\
$C_0= 0.0156$& Pin-SVM         & 68.80                          & 9.68                                       & -1.00                     \\
& C-SVM             & 67.09                          & 0.51                                       &-                      \\
\hline
Breast C.       & Unified Pin-SVM & 97.63                          & 0.95                                     & -0.25                     \\
$C_0= 0.0078$& Pin-SVM         & 97.63                          & 1.10                                      & 0.11                      \\
& C-SVM             & 85.80                          & 0.54                                       & -                     \\
\hline
Echo       & Unified Pin-SVM & 90.20                          & 0.04                                    & -0.51                     \\
$C_0= 0.0078$& Pin-SVM         & 74.51                          & 0.03                                      & 0.00                      \\
& C-SVM             & 74.51                          & 0.03                                       & -                     \\
\hline
Australian & Unified Pin-SVM & 87.24                          & 1.05                                      & -0.30                     \\
$C_0= 0.0313$& Pin-SVM         & 84.48                          & 0.64                                    & 0.00                      \\
& C-SVM             & 84.48                          & 0.63                                       & -                      \\
\hline
Bupa Liver & Unified Pin-SVM & 63.16                          & 0.19                                       & 0.00                      \\
$C_0= 0.0156$& Pin-SVM         & 63.16                          & 0.20                                       & 0.00                      \\
& C-SVM             & 63.16                          & 0.20                                    & -                      \\
\hline
Votes      & Unified Pin-SVM & 93.62                          & 0.29                                       & -0.08                     \\
$C_0= 0.0156$& Pin-SVM         & 94.47                          & 0.32                                       & -0.99                     \\
& C-SVM             & 85.11                          & 0.20                                       & -                      \\
\hline
Diabetes   & Unified Pin-SVM & 75.75                          & 1.72                                   & -0.59                     \\
$C_0= 0.0078$& Pin-SVM         & 67.91                          & 0.89                                       & 0.00                      \\
& C-SVM             & 67.91                          & 0.87                                       & -                      \\
\hline
Fertility  & Unified Pin-SVM & 94.00                          & 0.19                                      & -1.00                     \\
$C_0= 0.0078$& Pin-SVM         & 94.00                          & 0.01                                      & 0.00                      \\
&C-SVM             & 94.00                          & 0.01                                       & -                      \\
\hline
Sonar      & Unified Pin-SVM & 81.48                          & 0.07                                       & -0.63                     \\
$C_0= 0.0313$& Pin-SVM         & 77.78                          & 0.86                                       & -1.00                     \\
&C-SVM             & 75.93                          & 0.05                                      & -                      \\
\hline
Ecoil      & Unified Pin-SVM & 96.85                          & 0.15                                       & 0.00                      \\
$C_0= 2$& Pin-SVM         & 96.85                          & 0.15                                       & 0.00                      \\
& C-SVM             & 96.85                          & 0.15                                      & -                      \\
\hline
Parlx      & Unified Pin-SVM & 67.07                          & 0.75                                       & -1.00                     \\
$C_0= 0.0078$& Pin-SVM         & 67.07                          & 0.04                                       & -1.00                     \\
& C-SVM             & 67.07                          & 0.04                                       & -                      \\
\hline
Spambase     & Unified Pin-SVM &  68.39                      & 265.35                                       & -0.95                    \\
$C_0= 0.0078$& Pin-SVM         &  59.15                         & 68.12                                      & 0                     \\
& C-SVM             &           59.15                & 68.58                                      & _                      \\
\hline
\end{tabular}}
\label{tablelin}
\end{table}

\begin{table}[htp]
	\caption{Pin-SVM models with RBF kernel}
	{\scriptsize \begin{tabular}{|c|c|c|c|c|c|}\hline
		
		Dataset&     SVM models & 	Acc.   & 	Time(s)  & $\tau$ \\ \hline
		Monk1      & Unified Pin-SVM & 84.95                          & 0.22                                       & 0.12                      \\
		$p=1$ & Pin-SVM         & 84.95                          & 0.19                                     & 0.12                      \\
		$C_0=16$ & C-SVM             & 83.33                          & 0.17                                       & -                      \\
		\hline
		Monk2      & Unified Pin-SVM & 86.11                          & 0.28                                   & -0.32                     \\
	$p=0.5$ & Pin-SVM         & 85.65                         & 0.25                                     & 0                    \\
   $C_0=1$		& SVM             & 85.65                          & 0.25                                     & -                      \\
		\hline
		Monk3      & Unified Pin-SVM & 91.67                          & 0.17                                    & 0                      \\
		$p=2$, & Pin-SVM         & 91.67                          & 0.17                                      & 0                      \\
		$C_0=2$& C-SVM             & 91.67                          & 0.18                                      & -                     \\
		\hline
		Spect      & Unified Pin-SVM & 93.58                          & 0.06                                      & -0.59                     \\
	$p=0.0078$& Pin-SVM         & 93.58                          & 0.05                                     & 0                     \\
	$C_0=0.5$	&C-SVM             & 93.58                          & 0.05                                      & -                      \\
		\hline
		Pima       & Unified Pin-SVM & 76.07                          & 0.65                                       & 0.45                    \\
		$p= 0.5 $& Pin-SVM         & 76.07                          & 0.63                                       & 0.45                    \\
		$C_0= 0.0625$& C-SVM             & 75.85                          & 0.53                                       &-                      \\
		\hline
		German      & Unified Pin-SVM & 68.80                          & 0.95                                     & -0.14                     \\
		$p= 1$& Pin-SVM         & 68.00                          & 1.10                                      & 0                     \\
	$C_0= 2$	& C-SVM             & 68.00                          & 0.54                                       & -                     \\
		\hline
		Australian & Unified Pin-SVM & 87.59                          & 0.95                                      & -0.86                     \\
		$C_0= 2$ & Pin-SVM         & 82.41                          & 0.66                                    & 0.00                      \\
	$C_0= 0.0078$	& C-SVM             & 82.41                          & 0.66                                       & -                      \\
		\hline
		Bupa Liver & Unified Pin-SVM & 65.26                          & 0.34                                      & -0.74                      \\
	$p= 0.25$	& Pin-SVM         & 65.26                          & 0.32                                       & 0.84                     \\
	$C_0= 0.1250$	& C-SVM             & 64.21                          & 0.24                                    & -                      \\
		\hline
		Diabetes   & Unified Pin-SVM & 79.10                         & 1.80                                  & 0.01                    \\
		$p= 0.5$& Pin-SVM         & 79.10                          & 1.81                                       & 0.01                      \\
	$C_0= 0.0313$	& C-SVM             & 78.36                          & 0.91                                       & -                      \\
		\hline
	\end{tabular}}
\label{table2}
\end{table}

Table \ref{tablelin} lists the optimal performance of the existing $C$- SVM model, Pin-SVM model and Unified Pin-SVM model along with their training time and tunned parameters value. It can be observed that in the case of several datasets like Spect, Haberman, Echo, Australian, Diabetes, Sonar and Spambase, the use of proposed Unified Pin-SVM over existing Pin-SVM and C-SVM model can result in significant improvement of accuracy. It is because of the fact that unlike the existing Pin-SVM model, the proposed Unified Pin-SVM model also minimizes the pinball loss function for $ -1 \leq \tau < 0$ in true spirit. 

We repeat the similar numerical experiments with the existing $C$-SVM model, Pin-SVM model and proposed Unified Pin-SVM model for RBF kernel also. The numerical results are listed in Table \ref{table2}. Figure \ref{steady_state} shows the plot of  accuracy on several datasets obtained by the existing Pin-SVM model and proposed Unified Pin-SVM model against different $\tau$ values from the set $\{ -1,-0.9,....,..0.9,1\}$ with RBF kernel. These plots and numerical results are consistent with the observations which have been made in the linear kernel case. 

\section{Conclusions}
This paper proposes a significant improvement over the Pin-SVM model. For this, it re-look the pinball loss function for $ -1 \leq \tau < 0$ and its corresponding optimization problem used in the Pin-SVM model. It finds that the optimization problem used in (Huang et al, \cite{pinsvmpath}) fails to minimize the pinball loss function for $ -1 \leq \tau < 0$  in its true sense.  Thereafter, it develops the right optimization problem which can minimize the pinball loss function for $-1 \leq \tau < 0$ in its true sense.

It makes us realize that the Pin-SVM model requires to solve different QPP for its positive and negative $\tau$ values in $[-1,1]$. Taking motivation from this, we further propose a Unified Pin-SVM QPP which can be used to solve the Pin-SVM model without bothering the sign of its parameter $\tau$ in $[-1,1]$. The proposed Unified Pin-SVM model can obtain a significant improvement in accuracy over the Pin-SVM model, as it can also minimize the pinball loss function with $ -1 \leq \tau < 0$ in true sense. We have performed extensive numerical experiments with nineteen real-world datasets and shown empirically that the proposed Unified Pin-SVM model can always obtain an improvement over the existing Pin-SVM model.

\section*{Acknowledgments}
We are extremely grateful to the anonymous reviewers and Editor for their valuable comments that helped us to enormously improve the quality of the paper.

\bibliography{final}
\end{document}